\documentclass[10pt,twocolumn,letterpaper]{article}

\usepackage{iccv}              

\definecolor{iccvblue}{rgb}{0.21,0.49,0.74}
\usepackage[pagebackref,breaklinks,colorlinks,allcolors=iccvblue]{hyperref}

\usepackage[accsupp]{axessibility} 
\usepackage{wrapfig}
\usepackage{graphicx}
\usepackage{amsmath,amssymb}
\usepackage{booktabs}
\usepackage{times}
\usepackage{epsfig}
\usepackage{mmstyle}
\usepackage{adjustbox}
\usepackage{cuted}
\usepackage{mathtools}
\usepackage{mathabx}
\usepackage{bm}
\usepackage[boxed, ruled,linesnumbered]{algorithm2e}
\usepackage{algpseudocode}
\usepackage{tikz}
\usepackage{comment}
\usepackage{color}
\usepackage{multirow}
\usepackage{tabu}
\usepackage{makecell}

\newcommand{\myparagraph}[1]{\vspace{2pt}\noindent{\bf #1}}

\newcommand*\samethanks[1][\value{footnote}]{\footnotemark[#1]}

\def\paperID{9904} 
\def\confName{ICCV}
\def\confYear{2025}

\title{CATSplat: Context-Aware Transformer with Spatial Guidance\\for Generalizable 3D Gaussian Splatting from A Single-View Image}

\author{
Wonseok Roh\textsuperscript{\normalfont 1}\thanks{Equal contribution.} \quad
Hwanhee Jung\textsuperscript{\normalfont 1}\samethanks \quad
Jong Wook Kim\textsuperscript{\normalfont 1} \quad
Seunggwan Lee\textsuperscript{\normalfont 1}\\
Innfarn Yoo\textsuperscript{\normalfont 3} \quad
Andreas Lugmayr\textsuperscript{\normalfont 2} \quad
Seunggeun Chi\textsuperscript{\normalfont 4} \quad
Karthik Ramani\textsuperscript{\normalfont 4} \quad
Sangpil Kim\textsuperscript{\normalfont 1}\thanks{Corresponding author} \vspace{0.6em}\\
\textsuperscript{\normalfont 1}Korea University \quad
\textsuperscript{\normalfont 2}Google \quad 
\textsuperscript{\normalfont 3}CNAPS.AI Inc. \quad 
\textsuperscript{\normalfont 4}Purdue University
\vspace{-.3em}
}

\begin{document}
\maketitle
\begin{abstract}
Recently, generalizable feed-forward methods based on 3D Gaussian Splatting have gained significant attention for their potential to reconstruct 3D scenes using finite resources.
These approaches create a 3D radiance field, parameterized by per-pixel 3D Gaussian primitives, from just a few images in a single forward pass.
Unlike multi-view methods that benefit from cross-view correspondences, 3D scene reconstruction with a single-view image remains an underexplored area.
In this work, we introduce \textbf{CATSplat}, a novel generalizable transformer-based framework designed to break through the inherent constraints in monocular settings.
First, we propose leveraging textual guidance from a visual-language model to complement insufficient information from single-view image features.
By incorporating scene-specific contextual details from text embeddings through cross-attention, we pave the way for context-aware 3D scene reconstruction beyond relying solely on visual cues. 
Moreover, we advocate utilizing spatial guidance from 3D point features toward comprehensive geometric understanding under monocular settings.
With 3D priors, image features can capture rich structural insights for predicting 3D Gaussians without multi-view techniques.
Extensive experiments on large-scale datasets demonstrate the state-of-the-art performance of CATSplat in single-view 3D scene reconstruction with high-quality novel view synthesis.
\end{abstract}
\vspace{-2.6em}

\section{Introduction}
\label{sec:introduction}

3D scene reconstruction and novel view synthesis are fundamental tasks in modern computer vision and graphics, driving advancements across diverse domains~\cite{dalal2024gaussian, liu2024make, adamkiewicz2022vision, martin2021nerf}, such as virtual reality and autonomous navigation.
Together, they create 3D scene representations using 2D source images and produce realistic images from unseen perspectives.
Early approaches~\cite{mildenhall2021nerf, park2021nerfies, chen2022tensorf, barron2021mip} (\eg, NeRF) have made impressive progress through differentiable volume rendering.
However, they are still far from real-time scenarios due to the heavy computational demands. 
Unlike previous methods, 3D Gaussian Splatting (3DGS) based approaches~\cite{kerbl20233d, yu2024mip, yang2024deformable} have emerged as leading frontrunners, achieving high performance with real-time rendering capabilities.
They employ 3D Gaussians for explicit scene representations via efficient rasterization-based rendering.

\begin{figure}[t]
    \centering
    \vspace{-1.em}
    \includegraphics[width=\linewidth]{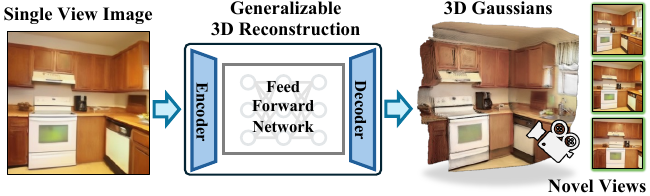}
    \vspace{-1.8em}
    \caption{
    Overview of the generalizable 3D scene reconstruction pipeline.
    The feed-forward network creates a 3D radiance field using 3D Gaussians, all within an end-to-end differentiable system.
    }
    \label{fig:outline}
    \vspace{-1.8em}
\end{figure}

Recently, generalizable feed-forward methods~\cite{charatan2024pixelsplat, chen2024mvsplat, zhang2024transplat, wewer2024latentsplat, szymanowicz2024flash3d} based on 3DGS~\cite{kerbl20233d} have attracted growing interest for their ability to reconstruct 3D scenes, even with limited resources like sparse view images.
They create a 3D radiance field parameterized by per-pixel Gaussian primitives from just a few input images (typically one or two) in a single forward pass without scene-specific optimization, as outlined in~\cref{fig:outline}.
For example, pixelSplat~\cite{charatan2024pixelsplat} samples Gaussian centers from probabilistic depth distributions using a multi-view epipolar geometry, while MVSplat~\cite{chen2024mvsplat} constructs cost volumes from source images to extract geometric cues.
Both approaches benefit from the cross-view correspondences between a pair of images to capture useful cues for the Gaussian parameters prediction.
In contrast to multi-view settings, monocular 3D scene reconstruction still remains challenging due to relatively constrained information, relying solely on an image.
Although Flash3D~\cite{szymanowicz2024flash3d} has pioneered a 3DGS-based generalizable single-view 3D scene reconstruction method with a pre-trained depth estimation network~\cite{piccinelli2024unidepth}, this area has yet to be fully explored. 

\begin{figure}[t]
    \centering
    \vspace{-.7em}
    \includegraphics[width=\linewidth]{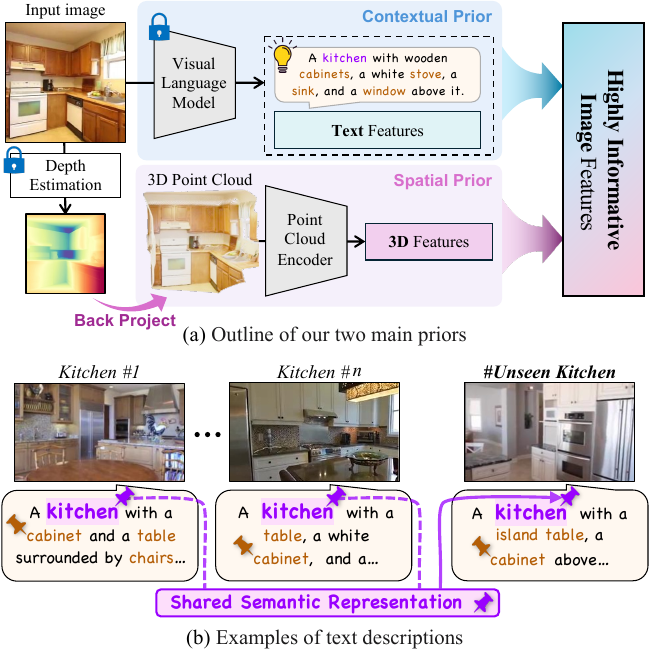}
    \vspace{-1.8em}
    \caption{
    We introduce \textbf{CATSplat}, a \textbf{C}ontext-\textbf{A}ware \textbf{T}ransformer with \textbf{S}patial Guidance for Generalizable 3D Gaussian Splatting from a single image.
    (a) Our two main priors, and
    (b) Examples of text descriptions (from the VLM) representing an input image.
    }
    \label{fig:intro}
    \vspace{-1.8em}
\end{figure}

To tackle the challenges in monocular scenarios, we introduce \textit{CATSplat}, a carefully designed transformer that leverages two intelligent guidance to supplement the insufficient information from single-view image features.
Based on the traditional generalizable 3DGS paradigm, which predicts Gaussian primitives from image features, we focus on enhancing these features with additional insights.
First, we propose using textual guidance as contextual priors.
One of the most promising ways to employ text guidance is through visual-language models (VLM)~\cite{liu2024visual, li2023blip, achiam2023gpt, zhu2023minigpt}.
They have showcased their potential in various vision tasks~\cite{koh2024generating, zhuang2024vlogger, huang2023language, lai2024lisa} to provide visual-linguistic knowledge learned from large-scale multimodal data.
Motivated by the success of VLMs, we utilize their text embeddings representing the input image to guide the network towards context-aware scene reconstruction, as in Fig.~\ref{fig:intro} (a).
Here, text features usually capture spatial context (e.g., \textit{kitchen}), object categories with their relationships (e.g., \textit{a table surrounded by chairs}), as well as overall compositions, as in Fig.~\ref{fig:intro} (b).
These text-embedded scene semantics not only enrich contextual understanding but also improve generalizability via shared semantic representations across similar yet distinct scenes.
For instance, when evaluating on an unseen \textit{kitchen} scene, the text embeddings serve as extra anchors, helping the network identify the scene as likely a \textit{kitchen}.
With explicit textual guidance, our network can better apply the knowledge learned from various \textit{kitchen} scenes (\textit{\#1}$\sim$$n$), boosting generalizability beyond solely relying on visual features.

In addition to the contextual guidance, we explore additional avenues to enhance the image features.
In generalizable tasks with sparse images, gaining insights into 3D geometrics is crucial to reconstruct robust scenes in 3D space.
Typically, multi-view methods~\cite{charatan2024pixelsplat, chen2024mvsplat} operate physical techniques like triangulation to capture rich 3D cues from cross-view perspectives.
However, such techniques are unavailable in monocular settings, leading to constrained geometric details.
In this context, we advocate for integrating 3D guidance into 2D features to enrich their spatial understanding.
Instead of merely using a 2D depth map from an off-the-shelf depth estimation model, as used in previous work~\cite{szymanowicz2024flash3d}, we further leverage its 3D representation as a backprojected point cloud.
Unlike depth maps, which are confined to a 2D grid, point clouds encode continuous x, y, and z-axis information in 3D space, capturing richer geometric structures.
Thus, we utilize point clouds as a source of structural guidance for reliable 3D perception.
Specifically, as in Fig.~\ref{fig:intro} (a), we extract 3D features from these points and strengthen image features through attention mechanisms.
Ultimately, our image features, influenced by two constructive priors, are highly informative for 3D Gaussian-based reconstruction.

Given landmark datasets, RealEstate10K (RE10K)~\cite{zhou2018stereo}, ACID~\cite{liu2021infinite}, KITTI~\cite{geiger2012we}, and NYUv2~\cite{silberman2012indoor}, we validate the generalizability and effectiveness of our novel framework.
To summarize, our main contributions are listed as follows:

\begin{itemize}
\item 
We introduce \textbf{CATSplat}, a novel generalizable framework for monocular 3D scene reconstruction.
We leverage the rich contextual cues of text embeddings from the VLM as insightful guidance toward context awareness, complementing limited information from a single image.
\item 
We propose 3D spatial guidance from 3D point features to enrich geometric details in single-view settings.
With 3D priors, image features can capture valuable cues for predicting 3D Gaussians without multi-view techniques.
\item We analyze the effectiveness of our method on challenging datasets. 
Extensive quantitative and qualitative experiments demonstrate that ours achieves new state-of-the-art performance on single-view 3D scene reconstruction. 
\end{itemize}

\section{Related Work}
\label{sec:related}

\begin{figure*}[t]
    \vspace{-1.5em}
    \centering
    \includegraphics[width=\linewidth]{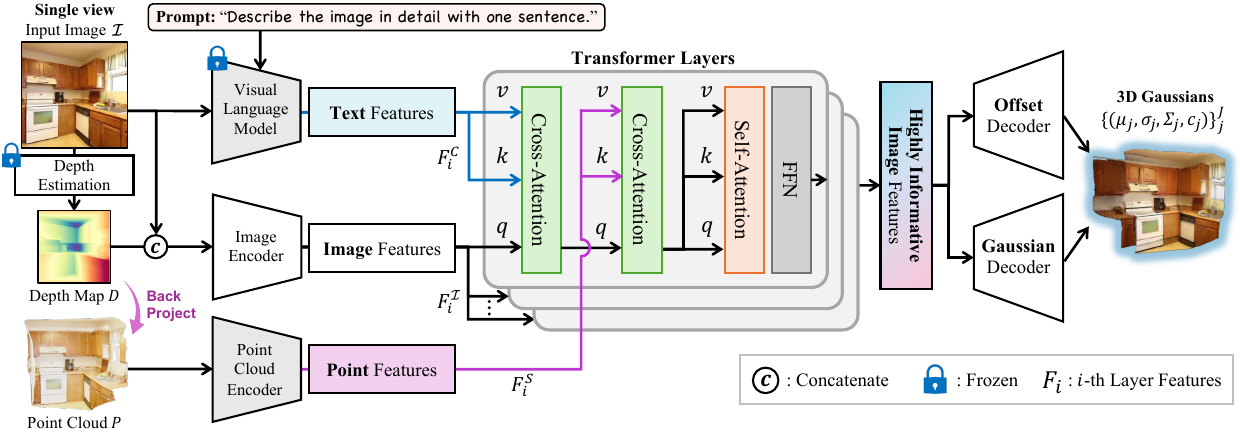}
    \vspace{-1.5em}
    \caption{
    Overview of \textbf{CATSplat} framework.
    CATSplat takes an image $\mathcal{I}$ and predicts 3D Gaussian primitives $\{(\bm{\mu}_j, \bm{\alpha}_j, \bm{\Sigma}_j, \bm{c}_j )\}^{J}_{j}$ to construct a scene-representative 3D radiance field in a single forward pass.
    Our primary goal is to go beyond the finite knowledge inherent in single-view image features leveraging our two innovative priors.
    Through cross-attention layers, we enhance image features $F^\mathcal{I}_i$ to be highly informative by incorporating valuable insights: contextual cues from text features $F^C_i$, and spatial cues from 3D point features $F^S_i$.
    }
    \label{fig:overview}
    \vspace{-1.em}
\end{figure*}

\myparagraph{Sparse-view 3D Reconstruction.}
Recent progress in neural fields~\cite{sitzmann2020implicit, xie2022neural, martin2021nerf} and volume rendering~\cite{lombardi2019neural, tagliasacchi2022volume} has advanced 3D reconstruction and novel view synthesis, even with sparse-view images.
FreeNeRF~\cite{yang2023freenerf} regularizes frequency to address few-shot neural rendering, while pixelNeRF~\cite{yu2021pixelnerf} predicts a neural radiance field in the camera coordinate using a feed-forward approach from few-view images.
More recently, 3D Gaussian Splatting (3DGS)~\cite{kerbl20233d} has revolutionized the field of 3D reconstruction, achieving real-time rendering.
Inspired by the success of 3DGS, pixelSplat~\cite{charatan2024pixelsplat} has pioneered the feed-forward network, which reconstructs a 3D radiance field parameterized by 3D Gaussian primitives from a pair of images.
Then, diverse multi-view generalizable 3DGS approaches~\cite{chen2024mvsplat, wewer2024latentsplat, zhang2024transplat} have since developed with a similar structure.
MVSplat~\cite{chen2024mvsplat} constructs cost volumes to capture cross-view similarities for accurate Gaussians, and latentSplat~\cite{wewer2024latentsplat} proposes variational Gaussians to encode uncertainty in a latent space.
Unlike these methods, monocular tasks remain more challenging due to limited details, without benefits of cross-view properties.

\myparagraph{Single-view 3D Reconstruction.}
Early approaches~\cite{wiles2020synsin, tucker2020single} have proposed practical strategies to overcome the constraints of single-view settings.
SynSin~\cite{wiles2020synsin} introduces a differentiable point renderer that projects a 3D point cloud from a single image into target views.
\citep{tucker2020single} predicts multiplane images (MPI)~\cite{zhou2018stereo} directly from a single image without correlations between multiple views.
In line with recent trends, single-view 3D reconstruction quality has significantly improved, thanks to innovations in NeRF~\cite{martin2021nerf} and 3DGS~\cite{kerbl20233d}.
Built upon NeRF, MINE~\cite{li2021mine} extends MPI to a continuous 3D representation, and BTS~\cite{wimbauer2023behind} predicts less complex continuous density fields from an image.
Recently, Splatter Image~\cite{szymanowicz2024splatter} involves 3DGS based on an image-to-image neural network for monocular object reconstruction. 
Also, Flash3D~\cite{szymanowicz2024flash3d} predicts pixel-wise Gaussian parameters for scene reconstruction in a single forward pass, using depth cues from a monocular depth estimation model~\cite{piccinelli2024unidepth}.
Based on the core paradigm of generalizable 3DGS frameworks, we proposes two beneficial priors to complement insufficient cues from an image, enhancing generalizability.

\myparagraph{Vision-Language Models for Vision Tasks.}
Visual Language Models (VLMs) have emerged as powerful tools for bridging the gap between visual and textual modalities~\cite{gan2022vision, long2022vision}, achieving outstanding performance in diverse vision tasks, such as image captioning~\cite{li2022blip, li2023blip, alayrac2022flamingo, yu2022coca, nguyen2024improving}, image-text retrieval~\cite{radford2021learning, jia2021scaling, lu2022cots, zhang2024user}, and visual question answering (VQA)~\cite{mishra2019ocr, qian2024nuscenes, hu2023promptcap}.
These models use large-scale image-text pair datasets to learn joint representations, encouraging seamless understanding and integration across modalities. 
Early approaches like CLIP~\cite{radford2021learning} and ALIGN~\cite{jia2021scaling} leverage contrastive learning to relate image and text data within a shared embedding space, enabling effective zero-shot generalization between them.
Recently, the success of Large Language Models (LLMs)~\cite{anil2023palm, touvron2023llama, chiang2023vicuna, brown2020language} has driven significant advancements in visual-language processing.
For example, BLIP-2~\cite{li2023blip} and LLaVA~\cite{liu2024visual} achieve strong performance in image captioning with context-rich visual descriptions based on LLMs~\cite{chiang2023vicuna, zhang2022opt, chung2024scaling}.
They aim to connect image features from visual encoders into the language space of pre-trained LLMs.
Motivated by the significance of VLMs, we employ contextual clues of their text embeddings to provide additional perspectives and enhance generalizability.
\section{Method}
\label{sec:method}

In this section, we introduce CATSplat, a novel generalizable framework for monocular 3D scene reconstruction with 3D Gaussian Splatting.
We first provide an overview of the whole pipeline (Sec.~\ref{sec:3-1} and Fig.~\ref{fig:overview}) and then elaborate on technical details: Context-Aware 3D Reconstruction (Sec.~\ref{sec:3-2}) and Spatial Guidance for 3D Insights (\cref{sec:3-3}).

\subsection{Overview}\label{sec:3-1}
Recent generalizable feed-forward frameworks~\cite{charatan2024pixelsplat, chen2024mvsplat, zhang2024transplat, wewer2024latentsplat, szymanowicz2024flash3d}  commonly follow a similar paradigm; they construct a 3D radiance field from $N$ sparse-view images $\mathcal{I}^{N} \in \mathbb{R}^{N \times H \times W \times 3}$ in a single forward pass with pixel-aligned $J$ Gaussian primitives $\{(\bm{\mu}_j, \bm{\alpha}_j, \bm{\Sigma}_j, \bm{c}_j )\}^{J}_{j}$, including position $\bm{\mu}_j$, opacity $\bm{\alpha}_j$, covariance $\bm{\Sigma}_j$, and spherical harmonics coefficients $\bm{c}_j$.
In this paradigm, it is challenging to reconstruct the vivid scene from a single image due to limited resources, comparing with multi-view configurations.
To overcome this constraint, we propose a carefully designed transformer that leverages two extra guidance for enhancing knowledge of single-view image features: (1) Textual Guidance, which provides explicit contextual clues for the scene, and (2) Spatial Guidance, which enriches three-dimensional structural information of 2D features, as illustrated in Fig.~\ref{fig:overview}.

\myparagraph{Feed-Forward Network with Transformer.}
From a single input image $\mathcal{I} \in \mathbb{R}^{H \times W \times 3}$, we first predict a depth map $D \in \mathbb{R}_{+}^{H \times W \times 1}$ as potential centers for Gaussians, employing a pre-trained monocular depth estimation model~\cite{piccinelli2024unidepth}.
Given $\mathcal{I}$ and its estimated depth map $D$, we channel-wise concatenate them as ${\mathcal{I}}^\prime \in \mathbb{R}^{H \times W \times 4}$, then feed $\mathcal{I}^\prime$ into a ResNet-based image encoder~\cite{he2016deep} to produce hierarchical depth-conditioned image features $F_{i}^{\mathcal{I}} \in \mathbb{R}^{H_{i} \times W_{i} \times D_{i}^{\mathcal{I}}}$.
Then, we utilize a multi-resolution transformer that encourages image features $F_{i}^{\mathcal{I}}$ to effectively represent both global structures and fine details across various resolutions, improving the overall understanding of the scene.
We specifically use three layers with three resolution features.
Based on transformer architecture, we extend the cross-attention mechanism to interact with our two novel priors, as described in Sec.~\ref{sec:3-2} and Sec.~\ref{sec:3-3}, further enriching the feature representation.
Through iterative layers, our transformer yields highly informative image features $\Tilde{F}_{i}^{\mathcal{I}} \in \mathbb{R}^{H_{i} \times W_{i} \times D_{i}^{\mathcal{I}}}$ well-suited for effective scene reconstruction in 3D space.
Finally, we estimate per-pixel 3D Gaussian parameters from $\Tilde{F}_{i}^{\mathcal{I}}$ using ResNet-based decoders, as detailed in Sec~\ref{sec:3-4}.


\subsection{Context-Aware 3D Reconstruction}\label{sec:3-2}
In real-world scenarios, diverse objects are usually placed in inconsistent patterns without following conventional rules.
These variabilities make monocular 3D scene reconstruction more challenging.
To extend the representational capability of image features, we propose textual information as a rich source of hidden context, enhancing generalizability.

\begin{figure}[t]
    \centering
    \vspace{-1.em}
    \includegraphics[width=\linewidth]{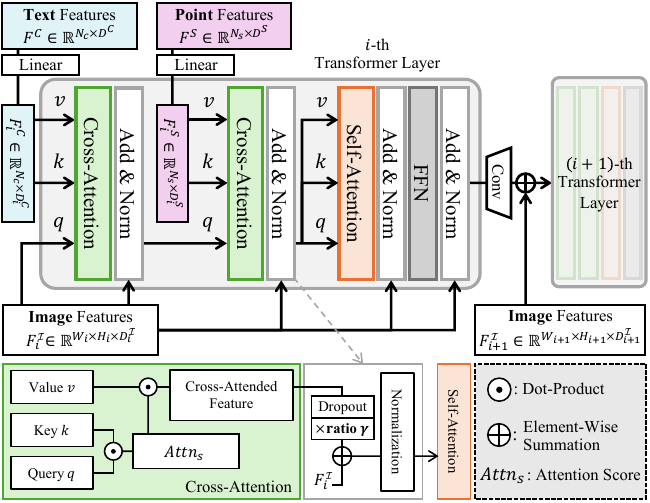}
    \vspace{-1.5em}
    \caption{
    Detailed transformer pipeline.
    In the $i$-th layer, we first operate cross-attention between $F_i^{\mathcal{I}}$ and $F_i^C$, then proceed cross-attention with $F_i^S$.
    We also use a ratio $\gamma$ to preserve visual information from $F_i^{\mathcal{I}}$ while incorporating extra cues from $F_i^C$ and $F_i^S$.
    }
    \label{fig:transformer}
    \vspace{-1.em}
\end{figure}

\myparagraph{Incorporation of Textual Cues.}
Recent advancements in large-scale visual language models~\cite{liu2024visual, li2023blip, achiam2023gpt, zhu2023minigpt} (VLM) have highlighted the benefits of their general embedded knowledge, which mirrors the diversity of real-world contexts. 
In this work, we take advantage of language-driven scene interpretations embedded in the text representations produced by these models.
With a single-view source image $\mathcal{I}$, we prompt the pre-trained VLM~\cite{liu2024visual} to generate a detailed, one-sentence description of the scene.
During this procedure, we utilize text embeddings $F^C \in \mathbb{R}^{N_c \times D^C}$ from a well-aligned multimodal space before they are processed into linguistic descriptions.
Our main focus is on the contextual details from $F^C$, such as object identities, spatial relationships, and scene semantics, which can potentially serve as influential biases for enhancing generalizability.
To softly incorporate supplemental cues from $F^C$ into image features $F^\mathcal{I}$, we employ iterative cross-attention layers.
For each transformer layer designed to use multi-scale features, we convert $F^C$ into $F_{i}^{C} \in \mathbb{R}^{N_c \times D^C_i}$ to align the dimension with its corresponding $F_{i}^{\mathcal{I}}$ using a linear layer, as illustrated in Fig.~\ref{fig:transformer}.
Given $F_{i}^{\mathcal{I}}$ and $F_{i}^{C}$, queries $\mathbf{Q}_i$ are projected from $F_{i}^{\mathcal{I}}$, and keys $\mathbf{K}_i$ and values $\mathbf{V}_i$ are from $F_{i}^{C}$, as follows:
\begin{equation}
    \mathbf{Q}_i = W_q \cdot F_{i}^{\mathcal{I}},\;\: \mathbf{K}_i = W_k \cdot F_{i}^{C},\;\: \mathbf{V}_i = W_v \cdot F_{i}^{C} 
\end{equation}
where $W$ denotes the learnable parameters of each projection layer.
Then, we associate them through cross-attention:
\vspace{-1.2em}
\begin{equation}
    F_i^{\mathcal{I}C} = Attn(\mathbf{Q}_i, \mathbf{K}_i, \mathbf{V}_i) = \text{Softmax}(\frac{\mathbf{Q}_i\cdot\mathbf{K}_i^T}{\sqrt{D_i}})\mathbf{V}_i
\label{eq:cross-ic}
\vspace{-.5em}
\end{equation}
where $F_i^{\mathcal{I}C}$ represents output features containing not only visual clues from $F_{i}^{\mathcal{I}}$ but also textual clues from $F_{i}^{C}$.
Finally, our iterative layers continuously establish valuable connections between source image features and additional contextual priors, facilitating more generalizable 3D reconstruction of real-world scenes under limited resources.


\subsection{Spatial Guidance for 3D Insights}\label{sec:3-3}
In multi-view configurations, each perspective contributes unique spatial information, boosting the reconstruction of complex three-dimensional structures.
Yet, single-view often falls short of 3D cues for comprehensive geometric understanding.
To bridge this gap, we introduce efficient spatial guidance based on the 3D representation of a 2D depth map, which provides a broader geometric context for reliable 3D perception independent of stereo vision expertise.

\myparagraph{Incorporation of Spatial Cues.}
Solid geometric awareness is essential for accurately depicting a scene within 3D space.
To capture 3D cues from a single image, traditional approaches~\cite{szymanowicz2024flash3d, szymanowicz2024splatter, li2021mine} often rely on depth information in a two-dimensional format.
Beyond its conventional use, we extend the estimated per-pixel 2D depth $d \in D$ into a full 3D representation for more direct spatial knowledge.
Given camera parameters $K$ = $diag(f_x, f_y, 1) \in \mathbb{R}^{3 \times 3}$, where $f$ denotes the focal length, we unproject $D$ into 3D space as point cloud $P \in \mathbb{R}^{H \times W \times 3}$, with each point $\bm{p} \in P$:
\vspace{-.2em}
\begin{equation}
    \bm{p} = K^{-1} \cdot \bm{u} \cdot d
    = (u_{x}d/f_x,u_{y}d/f_y,d)
\vspace{-.2em}
\end{equation}
where $\bm{u} = (u_x, u_y, 1) \in \mathcal{I}$ is one of the image pixels.
From this set of points $P$, we extract point features $F^S \in \mathbb{R}^{N_s \times D^S}$ using a PointNet-based encoder~\cite{qi2017pointnet} for better spatial reasoning.
These point embeddings usually encode important geometric details, from depth relationships to surface orientations, going beyond static depth information.
In order to integrate such valuable clues into image features while overcoming the domain gap between 2D and 3D representations, we leverage cross-attention layers.
Similar to the approach for textual cues, we project $F^S$ into $F_{i}^{S} \in \mathbb{R}^{N_s \times D^S_i}$ and further enrich context-guided image features $F^{\mathcal{I}C}_i$ (Eq.~\ref{eq:cross-ic}) from the previous cross-attention layer with $F_{i}^{S}$ as follows:
\vspace{-.5em}
\begin{equation}
\hspace{-.5em}
    F_i^{\mathcal{I}CS} = Attn(\mathbf{Q}^\prime_i, \mathbf{K}^\prime_i, \mathbf{V}^\prime_i) = \text{Softmax}(\frac{\mathbf{Q}^\prime_i\cdot{\mathbf{K}^\prime_i}^T}{\sqrt{D_i}})\mathbf{V}^\prime_i
\label{eq:cross-ics}
\vspace{-.3em}
\end{equation}
where $\mathbf{Q}^\prime_i$ are projected from $F_{i}^{\mathcal{I}C}$, and $\mathbf{K}^\prime_i$ and $\mathbf{V}^\prime_i$ are from $F_{i}^{S}$.
During the add and normalization process after cross-attention, as shown in Fig.~\ref{fig:transformer} below, we use the ratio $\gamma$ to preserve core visual information from the source image while incorporating practical cues from our two novel priors as:
\vspace{-.2em}
\begin{equation}
    \Tilde{F}_i^{\mathcal{I}CS} = \text{Norm}(F_{i}^{\mathcal{I}} + \bm{\gamma} \: \text{Dropout}(F_i^{\mathcal{I}CS}))
\label{eq:addnorm}
\vspace{-.2em}
\end{equation}
Then, we refine $\Tilde{F}_{i}^{\mathcal{I}CS}$ to $\Tilde{F}_{i}^{\mathcal{I}}$ with the self-attention layer, ensuring seamless knowledge enhancement across the feature space.
Ultimately, the final transformer output features $\Tilde{F}_{i}^{\mathcal{I}}$ are highly informative for robust 3D scene reconstruction, even with an image.
Note that we provide more technical details and the workflow algorithm in Supp (Sec.2). 

\subsection{Gaussian Parameters Prediction}\label{sec:3-4}
With insightful features $\Tilde{F}_{i}^{\mathcal{I}}$, we predict parameters for $J$ pixel-aligned 3D Gaussians $\{(\bm{\mu}_j, \bm{\alpha}_j, \bm{\Sigma}_j, \bm{c}_j )\}^{J}_{j}$ through ResNet-based decoders~\cite{he2016deep} to represent the 3D scene.

\myparagraph{Gaussian center $\bm{\mu}$.}
For precise scene reconstruction, we predict depth offsets $\delta \in \mathbb{R}_{+}^{H \times W \times 1}$ to refine per-pixel depth $d \in D$ and 3D offsets $\Delta_j \in \mathbb{R}^{3}$ for center-wise alignment following~\cite{szymanowicz2024splatter, szymanowicz2024flash3d}.
Then, we unproject the 2D refined depth $\Tilde{d} = d+\delta$ into 3D points using the provided camera parameters $K$ to produce potential centers.
Given $\Delta_j$ and projected points, the $j^{th}$ Gaussian center $\bm{\mu}_j$ is set as follows:
\vspace{-.3em}
\begin{flalign}
    \bm{\mu}_j &= K^{-1}\cdot\bm{u}\cdot\Tilde{d} + \Delta_j \\
    &= (u_{x}\Tilde{d}/f_x + \Delta_x,\;\:u_{y}\Tilde{d}/f_y + \Delta_y,\;\:\Tilde{d} + \Delta_z)
\vspace{-.4em}
\end{flalign}
where $\bm{u} = (u_x, u_y, 1) \in \mathcal{I}$ is one of the image pixels.

\myparagraph{Opacity $\bm{\alpha}$, Covariance $\bm{\Sigma}$, and Color $\bm{c}$.}
In line with previous generalizable feed-forward methods~\cite{charatan2024pixelsplat, chen2024mvsplat} using 3DGS, we operate convolutional layers to predict each parameter.
We use the sigmoid activation function for the opacity $\bm{\alpha}$ to ensure that values are bounded between 0 and 1.
Additionally, we estimate a rotation matrix $R$ and a scaling matrix $S$ to construct the covariance matrix $\bm{\Sigma} = RSS^{T}R^{T}$.
Also, for the color, we decode spherical harmonics coefficients $\bm{c}$. 

\myparagraph{Loss Function.}
Finally, we render images $\hat{\mathcal{I}}_t$ from novel viewpoints based on the reconstructed 3D scene using rasterization operation.
For training, we calculate the following loss $\mathcal{L}_{total}$ as the sum of the three losses to optimize the quality of the rendered images $\hat{\mathcal{I}}_t$ with GT target images $\mathcal{I}_t$:
\vspace{-.5em}
\begin{equation}
\mathcal{L}_{total} = \lambda_{\ell1}\mathcal{L}_{\ell1} + 
\lambda_{\text{ssim}}\mathcal{L}_{\textrm{ssim}} + \lambda_{\text{lpips}}\mathcal{L}_{\textrm{lpips}}
\label{eq:loss_mask}
\end{equation}
where $\mathcal{L}_{\textrm{ssim}}$ and $\mathcal{L}_{\textrm{lpips}}$ represent Structural Similarity Index (SSIM) and Learned Perceptual Image Patch Similarity (LPIPS)~\cite{zhang2018unreasonable} losses, respectively, and each $\lambda$ is a hyper-parameter to handle the strength of the respective loss term.

\section{Experiments}
\label{sec:experiments}

\subsection{Experimental Setup}
\vspace{-.4em}
\myparagraph{Datasets.}
In this study, we train and evaluate the overall performance using a large-scale dataset, RealEstate10K (RE10K)~\cite{zhou2018stereo}, containing home walkthrough videos.
We also use three additional datasets, NYUv2 (indoor)~\cite{silberman2012indoor}, ACID (nature)~\cite{liu2021infinite}, and KITTI (driving)~\cite{geiger2012we}, for cross-dataset experiments.
Detailed descriptions of datasets and implementation details are provided in the supplementary.


\myparagraph{Evaluation Metrics.}
We quantitatively evaluate the 3D reconstruction performance using three traditional metrics for novel view synthesis: PSNR, SSIM~\cite{wang2004image}, and LPIPS~\cite{zhang2018unreasonable}.
For comparisons with single-view 3D reconstruction methods, we evaluate three metrics on unseen target frames located 5 and 10 frames away from the input source image as well as a randomly sampled frame within a ±30 frame range, following the standard evaluation protocol of previous methods~\cite{li2021mine, szymanowicz2024flash3d}.
Also, to further evaluate our method, we adopt conventional interpolation and extrapolation protocols from pixelSplat~\cite{charatan2024pixelsplat} and latentSplat~\cite{wewer2024latentsplat}, respectively, following Flash3D~\cite{szymanowicz2024flash3d}.
For extrapolation, we sample target views up to 45 frames before or after the source frame.


\begin{table*}[t]
\vspace{-1.6em}
\small
\setlength{\tabcolsep}{6.8pt}
\begin{center}
\begin{tabu}{lccccccccccc}
\toprule
& \multicolumn{3}{c}{$n=5$ (frames)} & & \multicolumn{3}{c}{$n=10$ (frames)} & & \multicolumn{3}{c}{$n=\textit{Random}$ (frames)} \\
\cmidrule{2-4} \cmidrule{6-8} \cmidrule{10-12}
Method & PSNR $\uparrow$ & SSIM $\uparrow$ & LPIPS $\downarrow$ && PSNR $\uparrow$ & SSIM $\uparrow$ & LPIPS $\downarrow$ && PSNR $\uparrow$ & SSIM $\uparrow$ & LPIPS $\downarrow$ \\ \midrule
MPI \cite{tucker2020single} & 27.10 & 0.870 & -- && 24.40 & 0.812 & -- && 23.52 & 0.785 & -- \\
BTS \cite{wimbauer2023behind} & -- & -- & -- && -- & -- & -- && 24.00 & 0.755 & 0.194 \\
Splatter Image \cite{szymanowicz2024splatter} & 28.15 & 0.894 & 0.110 && 25.34 & 0.842 & 0.144 && 24.15 & 0.810 & 0.177 \\
MINE \cite{li2021mine} & 28.45 & 0.897 & 0.111 && 25.89 & 0.850 & 0.150 && 24.75 & 0.820 & 0.179 \\
Flash3D \cite{szymanowicz2024flash3d} & 28.46 & 0.899 & 0.100 && 25.94 & 0.857 & 0.133 && 24.93 & 0.833 & 0.160 \\
\midrule
CATSplat (Ours) & \textbf{29.09} & \textbf{0.907} & \textbf{0.094} && \textbf{26.44} & \textbf{0.866} & \textbf{0.125} && \textbf{25.45} & \textbf{0.841} & \textbf{0.151}  \\
\bottomrule
 \end{tabu}
\vspace{-0.7em}
\caption{
    Comparisons of Novel View Synthesis (NVS) performance with state-of-the-art \textbf{single-view} 3D reconstruction approaches on the RealEstate10K~\cite{zhou2018stereo} dataset.
    Following the standard protocol from~\cite{li2021mine, szymanowicz2024flash3d}, we evaluate NVS metrics on unseen target frames located $n$ frames away from the input source frame.
    Also, we randomly sample an extra target frame within 30 frames apart from the source frame.
}
\label{tab:single}
\end{center}
\end{table*}

\begin{table*}[t]
\small
\vspace{-1.6em}
\setlength{\tabcolsep}{8.7pt}
\begin{center}
\begin{tabu}{lccccccccc}
\toprule
& & & \multicolumn{3}{c}{RE10K Interpolation} & & \multicolumn{3}{c}{RE10K Extrapolation} \\
\cmidrule{4-6} \cmidrule{8-10}
Input & Method & Framework & PSNR $\uparrow$ & SSIM $\uparrow$ & LPIPS $\downarrow$ && PSNR $\uparrow$ & SSIM $\uparrow$ & LPIPS $\downarrow$ \\ \midrule
\multirow{5.}{*}{Two-View} & pixelNeRF \cite{yu2021pixelnerf} & NeRF & 20.51 & 0.592 & 0.550 && 20.05 & 0.575 & 0.567 \\
& Du \etal \cite{du2023learning} & NeRF & 24.78  & 0.820 & 0.213 && 21.83 & 0.790 & 0.242 \\
& pixelSplat \cite{charatan2024pixelsplat} & 3DGS & 26.09 & 0.864 & 0.136 && 21.84 & 0.777 & 0.216 \\
& latentSplat \cite{wewer2024latentsplat} & 3DGS & 23.93 & 0.812 & 0.164 && 22.62 & 0.777 & 0.196 \\
& MVSplat \cite{chen2024mvsplat} & 3DGS & \textbf{26.39} & \textbf{0.869} & \textbf{0.128} && \textbf{23.04} & \textbf{0.813} & \textbf{0.185} \\\midrule \\[-4.5mm] \midrule
\multirow{2.}{*}{Single-View} & Flash3D \cite{szymanowicz2024flash3d} & 3DGS & 23.87 & 0.811 & 0.185 && 24.10 & 0.815 & 0.185 \\
& CATSplat (Ours) & 3DGS & \textbf{25.23} & \textbf{0.835} & \textbf{0.159} && \textbf{25.35} & \textbf{0.837} & \textbf{0.159} \\
\bottomrule
\end{tabu}
\vspace{-0.7em}
\caption{
    Comparisons of NVS performance with state-of-the-art \textbf{few-view} 3D reconstruction approaches on the RealEstate10K~\cite{zhou2018stereo}.
    Although we mainly focus on comparing with the leading single-view method, Flash3D~\cite{szymanowicz2024flash3d}, we also provide scores of two-view methods for additional references.
    Following Flash3D, we use interpolation and extrapolation protocols from previous works, \cite{charatan2024pixelsplat} and \cite{wewer2024latentsplat}, respectively.
}
\label{tab:multi}
\end{center}
\vspace{-2.5em}
\end{table*}

\subsection{Performance Comparisons with SOTA Models}
\vspace{-.5em}
\myparagraph{Comparisons with Single-view Methods.}
In this section, we quantitatively compare our proposed framework CATSplat with existing state-of-the-art single-view 3D reconstruction methods~\cite{wiles2020synsin, tucker2020single, wimbauer2023behind, szymanowicz2024splatter, li2021mine, szymanowicz2024flash3d}.
Despite significant advancements through robust radiance field rendering techniques~\cite{martin2021nerf, kerbl20233d}, monocular 3D scene reconstruction has yet to be fully explored and still faces challenges under resource constraints.
To address this challenging task, we introduce a carefully designed transformer-based architecture with two novel priors, enriching image features to predict precise 3D Gaussians for scene representation.
As reported in Tab.~\ref{tab:single}, we evaluate novel view synthesis performance on the RealEstate10K~\cite{zhou2018stereo} dataset.
CATSplat consistently outperforms previous methods with new state-of-the-art scores in terms of PSNR, SSIM, and LPIPS across three target frame at distinct locations.
Specifically, CATSplat achieves high-quality rendering not only for nearby frames, such as those 5 or 10 frames apart, but also for frames randomly located at far distances (within a ±30 frame range).
These results demonstrate that our novel priors effectively complement limited cues from ResNet-based single image features.

\myparagraph{Interpolation and Extrapolation.}
In multi-view setups, novel view synthesis is typically evaluated on target frames within the range of multiple input images (interpolation) and outside their range (extrapolation).
In Tab.~\ref{tab:multi}, to further validate our method, we evaluate CATSplat across both conventional settings, 
as established in Flash3D~\cite{szymanowicz2024flash3d}, a prominent single-view 3D scene reconstruction model.
While our primary focus is on comparing with Flash3D, we also provide scores of multi-view methods~\cite{yu2021pixelnerf, du2023learning, charatan2024pixelsplat, wewer2024latentsplat, chen2024mvsplat} for additional references.
First, CATSplat significantly surpasses Flash3D in the interpolation setup.
Although our results are somewhat lower than recent two-view methods, which are robust for intermediate views using cross-view correspondence, ours achieves competitive scores.
Moreover, for the extrapolation setup, CATSplat outperforms Flash3D by large margins.
Notably, these impressive scores even exceed previous two-view methods despite using only a single image.
In such extrapolation setups, target frames are usually over 45 frames away from the source image, representing nearly unseen views.
These findings highlight the effectiveness of our novel priors, providing valuable insights for handling distant target views.
Specifically, textual anchors based on the shared semantic representation of the scenes, along with deep spatial understanding from point features encoding continuous x, y, and z-axis knowledge, enhance generalizability in relatively restrictive monocular settings.


\begin{table}[t]
\begin{center}
\vspace{.1em}
\resizebox{\linewidth}{!}{%
\begin{tabular}{@{}ccccc@{}} \toprule
    Cross Dataset & Method & PSNR $\uparrow$ & SSIM $\uparrow$ & LPIPS $\downarrow$\\\midrule
    \multirow{2.3}{*}{\makecell{RE10K \\ $\rightarrow$ NYUv2}} & Flash3D~\cite{szymanowicz2024flash3d} & 25.09 & 0.775 & 0.182 \\
    & CATSplat (Ours) & \textbf{25.57} & \textbf{0.781} & \textbf{0.157} \\\midrule
    \multirow{2.3}{*}{\makecell{RE10K \\ $\rightarrow$ ACID}} & Flash3D~\cite{szymanowicz2024flash3d} & 24.28 & 0.730 & 0.263 \\
    & CATSplat (Ours) & \textbf{24.73} & \textbf{0.739} & \textbf{0.250} \\\midrule
    \multirow{2.3}{*}{\makecell{RE10K \\ $\rightarrow$ KITTI}} & Flash3D~\cite{szymanowicz2024flash3d} & 21.96 & 0.826 & 0.132 \\
    & CATSplat (Ours) & \textbf{22.43} & \textbf{0.833} & \textbf{0.122} \\
\bottomrule
\end{tabular}}
\vspace{-0.7em}
\caption{
Comparisons of cross-dataset generalization with the state-of-the-art single-view 3DGS method, Flash3D~\cite{szymanowicz2024flash3d}, on various real-world datasets: NYUv2~\cite{silberman2012indoor}, ACID~\cite{liu2021infinite}, and KITTI~\cite{geiger2012we}.
}
\vspace{-2.9em}
\label{tab:cross}
\end{center}
\end{table}

\begin{figure*}[t]
    \vspace{-2.em}
    \centering
    \includegraphics[width=\linewidth]{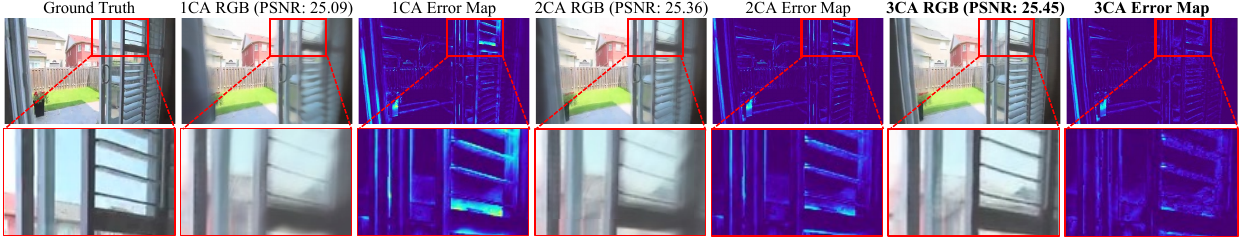}
    \vspace{-1.9em}
    \caption{
    Ablation study to see the effect of iteratively incorporating our novel priors on the RE10K~\cite{zhou2018stereo} ($n$=\textit{Random}).
    For clear ablations, we keep the number of entire transformer layers consistent across the experiments and adjust only the number of cross-attentions (CA).
    }
    \label{fig:error}
    \vspace{-1.4em}
\end{figure*}

\vspace{-.2em}
\myparagraph{Cross-dataset Generalization.}
In Tab.~\ref{tab:cross}, we demonstrate the strong generalizability of CATSplat across three different cross-dataset settings.
In each case, we train our model on RE10K~\cite{zhou2018stereo} and directly test it on the target datasets in a zero-shot manner.
We first evaluate the generalization on the NYUv2~\cite{silberman2012indoor}, which contains indoor scenes similar to the RE10K.
CATSplat adeptly synthesizes images for previously unseen indoor environments.
Then, we focus on outdoor scenarios with more significant domain gaps; specifically, the ACID~\cite{liu2021infinite} includes nature landscapes captured by aerial drones, and KITTI~\cite{geiger2012we} comprises driving scenes tailored for autonomous driving.
Within these challenging conditions, where filming techniques or object types (\eg, \textit{cars, buildings}) are dissimilar, CATSplat showcases superior generalizability than the latest method, Flash3D~\cite{szymanowicz2024flash3d}.
Through a series of experiments, we confirm the power of our intelligent priors in improving the informativeness of single-view image features for dynamic real-world scene reconstruction.
We thoroughly analyze this in Supp (Sec.4.1).

\begin{table}[t]
\begin{center}
\resizebox{\linewidth}{!}{%
\begin{tabular}{@{}lcccccc@{}} \toprule
& \multicolumn{3}{c}{$n=\text{10}$ (frames)} & \multicolumn{3}{c}{$n=\textit{Random}$ (frames)} \\\cmidrule{2-4} \cmidrule{5-7}
    Method & PSNR $\uparrow$ & SSIM $\uparrow$ & LPIPS $\downarrow$ & PSNR $\uparrow$ & SSIM $\uparrow$ & LPIPS $\downarrow$\\\midrule
    Baseline & 26.04 & 0.857 & 0.132 & 25.02 & 0.834 & 0.159\\
    w/ Contextual & 26.40 & 0.864 & 0.127 & 25.40 & 0.838 & 0.153\\
    w/ Spatial & 26.38 & 0.864 & 0.127 & 25.42 & 0.837 & 0.153\\
    CATSplat & \textbf{26.44} & \textbf{0.866} & \textbf{0.125} & \textbf{25.45} & \textbf{0.841} & \textbf{0.151}\\
\bottomrule
\end{tabular}}
\vspace{-0.7em}
\caption{
Ablation study to explore the effect of our two intelligent priors (Contextual and Spatial) across three different settings, as in Tab.~\ref{tab:single}, on the RE10K~\cite{zhou2018stereo} dataset.
Here, the ``Baseline'' indicates our basic transformer architecture without any proposed priors.
}
\vspace{-2.3em}
\label{tab:ablation}
\end{center}
\end{table}

\subsection{Ablation Studies}
\vspace{-.3em}
\myparagraph{Effect of Contextual and Spatial Priors.}
In Tab.~\ref{tab:ablation}, we evaluate variants of our method with/ and w/o Contextual and Spatial priors.
Here, the Baseline refers to our basic multi-resolution transformer architecture, excluding cross-attention with any of our proposed priors.
The addition of each prior consistently enhances the visual quality of the rendered images from target novel perspectives. 
With contextual priors, the improvements across all metrics underscore the significance of incorporating extra context details for effective scene reconstruction.
Also, spatial priors contribute impressive gains within all target settings, providing a more extensive geometric context for 3D 
understanding.
Ultimately, combining both valuable priors together leads to further advancements, achieving the best scores.
These results highlight that each prior plays a meaningful role in complementing limited details from single-view features.

\vspace{-.1em}
\myparagraph{Iteratively Incorporating Priors.}
Based on transformer, our feed-forward network seamlessly integrates two novel priors through iterative cross-attention layers.
In Fig.~\ref{fig:error}, we explore the effect of varying the number of cross-attention iterations using rendered images with corresponding error maps.
Specifically, we keep the total layers of the transformer consistent at three and apply cross-attention either in the first layer only, across two layers, or throughout all three layers.
Across experiments, increasing cross-attention iterations leads to more precise, less blurry image synthesis with fewer errors.
These improvements in visual quality through iterative incorporation underline the potential of our priors.

\vspace{-.1em}
\myparagraph{Vs. Pre-trained Visual Priors.}
Our primary goal in incorporating our two priors is to utilize comprehensive scene details that a simple image encoder may fail to capture from an image.
In Tab.~\ref{tab:visual_priors}, we validate the practical benefits of our two priors by replacing them in the cross-attention layers with visual priors from pre-trained image encoders, such as DINOv2~\cite{oquab2023dinov2} and ConvNeXt V2~\cite{woo2023convnext}, on large-scale image datasets.
The advanced image processing intelligence of these models, developed from extensive real-world images, contributes to moderate gains.
However, since they mainly focus on visible aspects of the scene and our model is inherently designed to handle image data, injecting additional visual cues may introduce redundancy rather than meaningful benefits.
In contrast, VLM extends beyond pure image processing by integrating language-driven insights, providing extra markers for our network to address unseen scenes.
Also, with point features, our network captures more global geometric properties than the localized features from image encoders, enabling more accurate 3D Gaussian predictions.
By leveraging both priors, we overcome the straightforward dependence on visual patterns, achieving notable increases.

\begin{figure*}[t]
    \vspace{-2.em}
    \centering
    \includegraphics[width=\linewidth]{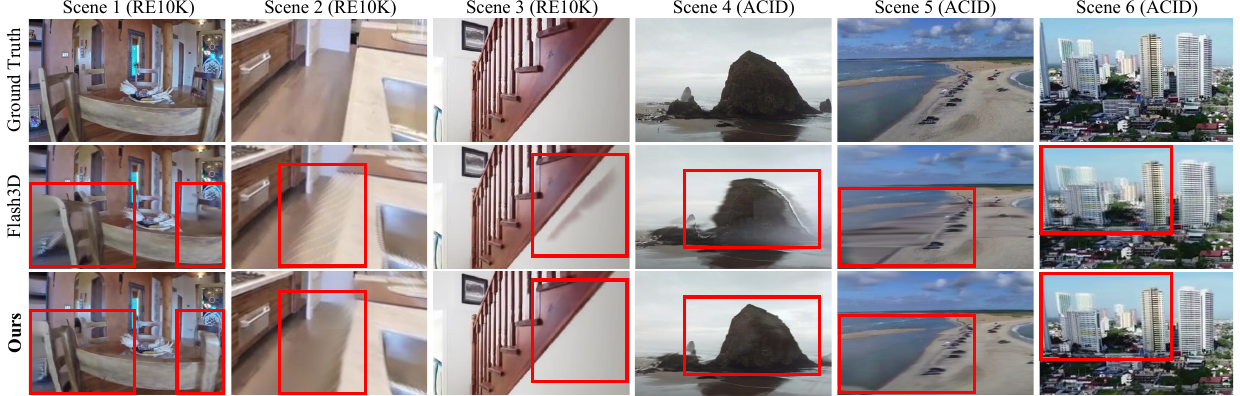}
    \vspace{-2.em}
    \caption{
    Qualitative comparisons of NVS performance between Flash3D~\cite{szymanowicz2024flash3d} and ours with Ground Truth on the novel view frames from RealEstate10K~\cite{zhou2018stereo} and ACID~\cite{liu2021infinite} (cross-dataset).
    We provide more visual results and details of user study in the supplementary material.
    }
    \label{fig:qualitative}
    \vspace{-1.5em}
\end{figure*}

\begin{table}[t]
\begin{center}
\vspace{0.05em}
\resizebox{\linewidth}{!}{%
\begin{tabular}{@{}lcccccc@{}} \toprule
& \multicolumn{3}{c}{$n=\text{10}$ (frames)} & \multicolumn{3}{c}{$n=\textit{Random}$ (frames)} \\\cmidrule{2-4} \cmidrule{5-7}
    Method & PSNR $\uparrow$ & SSIM $\uparrow$ & LPIPS $\downarrow$ & PSNR $\uparrow$ & SSIM $\uparrow$ & LPIPS $\downarrow$\\\midrule
    w/ ConvNeXt-B & 26.15 & 0.856 & 0.132 & 25.09 & 0.832 & 0.158\\
    w/ ConvNeXt-L & 26.17 & 0.857 & 0.132 & 25.12 & 0.833 & 0.157\\
    w/ DINOv2-B & 26.17 & 0.858 & 0.131 & 25.11 & 0.833 & 0.157\\
    w/ DINOv2-g & 26.19 & 0.859 & 0.131 & 25.17 & 0.834 & 0.156\\
    \midrule
    w/ Contextual & \textbf{26.40} & \textbf{0.864} & \textbf{0.127} & 25.40 & \textbf{0.838} & \textbf{0.153}\\
    w/ Spatial & 26.38 & \textbf{0.864} & \textbf{0.127} & \textbf{25.42} & 0.837 & \textbf{0.153}\\
\bottomrule
\end{tabular}}
\vspace{-0.7em}
\caption{
Ablation study to see the effect of our novel priors versus visual priors from large-scale pre-trained image encoders, such as DINOv2~\cite{oquab2023dinov2} and ConvNeXt V2~\cite{woo2023convnext}, on the RE10K~\cite{zhou2018stereo} dataset.
}
\vspace{-1.6em}
\label{tab:visual_priors}
\end{center}
\end{table}

\begin{table}[t]
\begin{center}
\resizebox{\linewidth}{!}{%
\begin{tabular}{@{}lcccccc@{}} \toprule
& \multicolumn{3}{c}{$n=\text{10}$ (frames)} & \multicolumn{3}{c}{$n=\textit{Random}$ (frames)} \\\cmidrule{2-4} \cmidrule{5-7}
    Method & PSNR $\uparrow$ & SSIM $\uparrow$ & LPIPS $\downarrow$ & PSNR $\uparrow$ & SSIM $\uparrow$ & LPIPS $\downarrow$\\\midrule
    Baseline & 26.04 & 0.857 & 0.132 & 25.02 & 0.834 & 0.159\\
    w/ Scene Type & 26.14 & 0.859 & 0.130 & 25.13 & 0.835 & 0.158\\
    w/ Object List & 26.23 & 0.862 & 0.128 & 25.25 & 0.836 & 0.155\\
    w/ Extended & 26.31 & 0.862 & 0.128 & 25.29 & 0.837 & 0.154\\
    w/ Single Sent. & \textbf{26.40} & \textbf{0.864} & \textbf{0.127} & \textbf{25.40} & \textbf{0.838} & \textbf{0.153}\\
\bottomrule
\end{tabular}}
\vspace{-0.7em}
\caption{
Ablation study to see the impact of different text description formats on generalizable tasks.
The ``Baseline'' is as in Tab.~\ref{tab:ablation}.
}
\vspace{-2.5em}
\label{tab:ablation_text}
\end{center}
\end{table}

\myparagraph{Analysis of Context Details.}
In Tab.~\ref{tab:ablation_text}, we explore how different context details embedded in text features from pre-trained VLM~\cite{liu2024visual} influence generalizability.
We conduct experiments with four different prompt styles: identifying the scene type (\eg, \textit{bedroom}), listing objects (\eg, \textit{lamp, bed}), describing the scene with a detailed single sentence, and two or more sentences.
While scene type or object list provides certain clues, their impact on performance is relatively modest.
In contrast, sentence-level text embeddings contain more practical context details, such as texture, object relations, and overall composition, for enhancing generalizability.
But extended versions may introduce overstatements, potentially confusing the network.
Thus, we use single sentence embeddings that offer proper yet unexaggerated details.
We further discuss this in Supp (Sec.4.2, 4.3).

\myparagraph{Analysis of Geometric Cues.}
To capture geometric cues under limited resources, it is crucial to guide the network with practical spatial information.
In Tab.~\ref{tab:ablation_3d}, we examine strategies to enhance geometrical knowledge from a single image.
Our base transformer network, called Baseline, concatenates depths with an image to extract depth-conditioned features.
We first evaluate using only the image, excluding depth concatenation, and observe drops in overall scores.
This highlights the meaningful role of the geometric condition.
Then, we replace the depth concatenation in the Baseline with unprojected 3D point concatenation.
While using 3D points yields slight gains, there is no significant benefit over depth.
Beyond simple concatenation, we employ attention strategies to integrate geometric cues seamlessly.
We finally observe that cross-attention with point features greatly contributes to comprehensive 3D understanding.
These validate the efficacy of our 3D spatial guidance incorporation.
\vspace{-.4em}

\begin{table}[t]
\begin{center}
\resizebox{\linewidth}{!}{%
\begin{tabular}{@{}lcccccc@{}} \toprule
& \multicolumn{3}{c}{$n=\text{10}$ (frames)} & \multicolumn{3}{c}{$n=\textit{Random}$ (frames)} \\\cmidrule{2-4} \cmidrule{5-7}
    Method & PSNR $\uparrow$ & SSIM $\uparrow$ & LPIPS $\downarrow$ & PSNR $\uparrow$ & SSIM $\uparrow$ & LPIPS $\downarrow$\\\midrule
    Baseline & 26.04 & 0.857 & 0.132 & 25.02 & 0.834 & 0.159\\
    w/o Depth Conc.\hspace{-.9em} & 25.91 & 0.855 & 0.134 & 24.82 & 0.827 & 0.165\\
    w/ Point Conc. & 26.06 & 0.857 & 0.132 & 25.04 & 0.834 & 0.158\\
    w/ Depth Feat. & 26.18 & 0.859 & 0.130 & 25.16 & 0.835 & 0.157\\
    w/ Point Feat. & \textbf{26.38} & \textbf{0.864} & \textbf{0.127} & \textbf{25.42} & \textbf{0.837} & \textbf{0.153}\\
\bottomrule
\end{tabular}}
\vspace{-0.7em}
\caption{
Ablation study to explore strategies for enriching geometric knowledge from a single image.
The ``Baseline'' is as in \cref{tab:ablation}.
}
\vspace{-2.7em}
\label{tab:ablation_3d}
\end{center}
\end{table}

\vspace{-1em}
\subsection{Visual Comparisons}
\vspace{-.5em}
\myparagraph{Qualitative Analysis.}
In Fig.~\ref{fig:qualitative}, we qualitatively compare rendered images from ours and Flash3D~\cite{szymanowicz2024flash3d}, along with ground truth for solid comparisons.
In Scene 1 (\textit{chair}) and 2 (\textit{sink}), ours achieves more precise object placement with less blurriness compared to Flash3D.
Also, in Scene 3 (\textit{stair}), CATSplat clearly represents a low-texture area, whereas Flash3D struggles with blotchy artifacts. 
Moreover, ours outperforms Flash3D in cross-dataset scenarios.
In Scene 4 and 5, ours captures well-defined edges; in Scene 6, ours renders a more detailed image from an aerial view of the complex cityscape.
In addition to comparing rendered RGBs, we assess the quality of 3D Gaussians with corresponding depth maps (Supp Sec.4.6).
These findings confirm the significance of our priors for novel view synthesis.

\myparagraph{User Study.}
In Tab.~\ref{tab:user}, we validate our method through human evaluation.
We randomly selected 60 and 20 scenes from the RE10K~\cite{zhou2018stereo} and ACID~\cite{liu2021infinite} datasets, and recruited 100 participants via Amazon Mechanical Turk.
We present two types of questions with rendered images: (i) preferring between ours and Flash3D~\cite{szymanowicz2024flash3d} based on performance, and (ii) rating the visual quality on a 7-point Likert scale.
For all evaluations, ours strongly outperforms Flash3D by a significant margin across both datasets.
Also, the narrow confidence interval highlights the consistency of these results.


\begin{table}[h]
\begin{center}
\vspace{-.8em}
\resizebox{\linewidth}{!}{%
\begin{tabular}{@{}lccccc@{}} \toprule
& \multicolumn{2}{c}{RE10K~\cite{zhou2018stereo}} && \multicolumn{2}{c}{ACID~\cite{liu2021infinite}} \\\cmidrule{2-3} \cmidrule{5-6}
    Method & Preference ($\%$) & Likert $\uparrow$ && Preference ($\%$) & Likert $\uparrow$ \\\midrule
    Flash3D~\cite{szymanowicz2024flash3d} & 11.58$\pm$1.09 & 4.56$\pm$0.30 && 8.59$\pm$0.63 & 4.14$\pm$0.21 \\
    CATSplat (Ours) & \textbf{88.42$\pm$1.09} & \textbf{6.04$\pm$0.22} && \textbf{91.41$\pm$0.63} & \textbf{5.27$\pm$0.18} \\\bottomrule
\end{tabular}}
\vspace{-0.7em}
\caption{
User study comparisons. We report mean preference percentage and a 7-point Likert scale with a 95\% confidence interval.
}
\vspace{-2.7em}
\label{tab:user}
\end{center}
\end{table}

\section{Conclusion}
\vspace{-.2em}
\label{sec:conclusion}
We introduce CATSplat, a novel generalizable 3DGS frame work using a single-view image.
Our core objective is to transcend the constraints of relying on a single image.
To this end, we propose two priors: (i) contextual priors from VLM text embeddings towards context-aware 3D scene reconstruction, and (ii) spatial priors from 3D point features for comprehensive geometric understanding.
Extensive experiments demonstrate the superiority of CATSplat.
While our method excels in monocular 3D scene reconstruction, ours might be less effective in occluded or truncated areas.
Besides, our current training relies on the RealEstate10K dataset; however, with diverse large-scale datasets, CATSplat would be more suitable for real-world applications.

\section*{Acknowledgement}
This work was supported by Culture, Sports and Tourism R\&D Program through the Korea Creative Content Agency grant funded by the Ministry of Culture, Sports and Tourism 
(International Collaborative Research and Global Talent Development for the Development of Copyright Management and Protection Technologies for Generative AI, RS-2024-00345025, 38\%;
Research on neural watermark technology for copyright protection of generative AI 3D content, RS-2024-00348469, 25\%;
Development of sketch-based semantic 3D modeling technology for creating user-centric Metaverse content spaces for indoor spaces, RS-2023-00227409, 10\%), 
the National Research Foundation of Korea (NRF) grant funded by the Korea government (MSIT) (RS-2025-00521602, 25\%),
Institute of Information \& communications Technology Planning \& Evaluation (IITP) \& ITRC (Information Technology Research Center) grant funded by the Korea government (MSIT) (No.RS-2019-II190079, Artificial Intelligence Graduate School Program (Korea University), 1\%; 
IITP-2025-RS-2024-00436857, 1\%), 
and
Artificial intelligence industrial convergence cluster development project funded by the Ministry of Science and ICT (MSIT, Korea) \& Gwangju Metropolitan City.

{
    \small
    \bibliographystyle{ieeenat_fullname}
    \bibliography{main}
}

\end{document}


\maketitlesupplementary

\section*{Overview}
In this supplementary material, we provide further explanations and visualizations of our main paper, \textit{``CATSplat: Context-Aware Transformer with Spatial Guidance for Generalizable 3D Gaussian Splatting from A Single-View Image''}.
First, we elaborate on the specifics of our user study (Sec.~\ref{supp:userstudy}). 
Then, we present additional technical details on the CATSplat architecture (Sec.~\ref{supp:architecture}). 
Also, we describe the implementation and datasets in more detail (Sec.~\ref{supp:setup}).
Moreover, we provide more quantitative and qualitative experimental results to further validate the robustness of CATSplat for 3D reconstruction and novel view synthesis (Sec.~\ref{supp:experiments}).
We finally discuss the limitations and future directions (Sec.~\ref{supp:limitations}).
\section{User Study Details}
\label{supp:userstudy}

\begin{figure}[h]
    \centering
    \includegraphics[width=\linewidth]{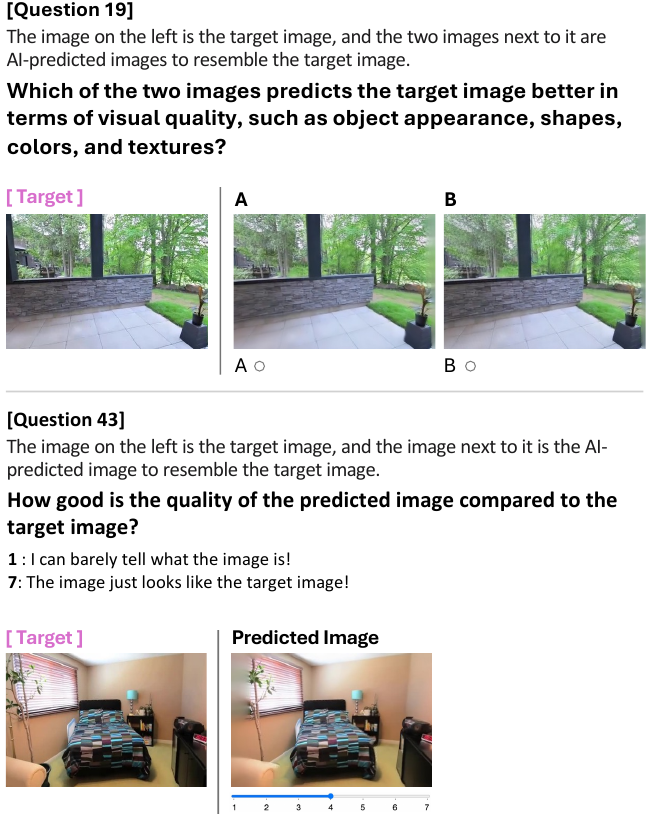}
    \vspace{-1.5em}
    \caption{
    Examples of two types of user study questions.
    The first type of question (above) asks about preference between ours and Flash3D~\cite{szymanowicz2024flash3d}, and the second (below) requires participants to rate the visual quality of the rendered image compared to the target.
    }
    \label{fig:userstudy}
    \vspace{-1.5em}
\end{figure}

We conduct a user study to validate our method from the perspective of human perception, as described in Sec.4.4 in the main paper.
Through Amazon Mechanical Turk (AMT), a widely used platform for user studies, we recruited 100 participants.
We randomly sample 60 scenes from the RE10K~\cite{zhou2018stereo} evaluation set and 20 from the ACID~\cite{liu2021infinite} evaluation set.
Then, we use rendered images from these sampled scenes for the survey questions.
With rendered images and corresponding ground truth target images, we ask two types of questions, as shown in Fig.~\ref{fig:userstudy}.
For the first type of question, we show two rendered images, one from CATSplat and the other from Flash3D~\cite{szymanowicz2024flash3d}, along with a target image, and ask, \textit{``Which of the two images predicts the target image better in terms of visual quality, such as object appearance, shapes, colors, and textures?''}.
For the second type of question, we request participants to rate the visual quality of the rendered image from either method (CATSplat or Flash3D) on a 7-point Likert scale, with the question, \textit{``How good is the quality of the rendered image compared to the target image?''}.
We also include control questions to verify the 
reliability of responses from each participant by displaying the ground truth image as the rendered image and asking participants to rate it based on the same ground truth image, where the results are expected to be obviously high.
Moreover, the method names are anonymized and presented in random order to minimize bias.
We finally gathered 9,000 responses on RE10K and 6,000 responses on ACID (\ie, 30 questions for type one and 30 rating questions for each CATSplat and Flash3D on RE10K, as well as 20 questions for type one and 20 rating questions for each on ACID).
Given responses from all participants, we report scores with 95\% confidence intervals, as shown in Tab.7 of the main paper.
Specifically, for the first type of question, which requires participants to choose between two rendered images, we utilize a binomial proportion confidence interval to analyze preferences.
In the case of the second type, which queries to rate the visual quality of a single rendered image, we use a normal distribution confidence interval to analyze the average rating score.
Ultimately, the results underscore the superiority of our method, as CATSplat is notably preferred and receives higher ratings compared to the latest method.

\section{Architecture Details}
\label{supp:architecture}

\begin{figure*}[t]
    \vspace{-1.5em}
    \centering
    \includegraphics[width=\linewidth]{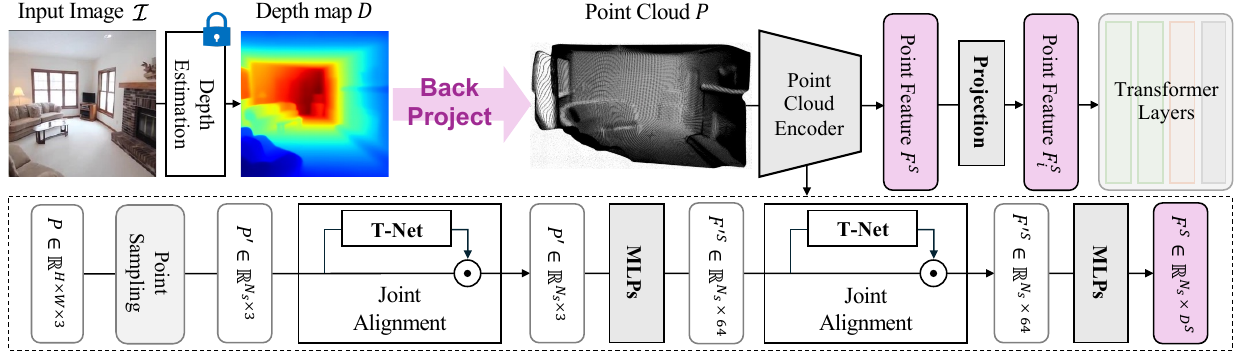}
    \vspace{-1.5em}
    \caption{
    Detailed architecture of 3D point feature extraction from a monocular input image $\mathcal{I}$.
    Our point cloud encoder takes back-projected points $P$ and produces point features $F^S$ based on the PointNet~\cite{qi2017pointnet} structure.
    Here, T-Net indicates an affine transform network.
    }
    \label{fig:3d_procedure}
    \vspace{-1.em}
\end{figure*}

\begin{figure}[t]
    \centering
    \includegraphics[width=\linewidth]{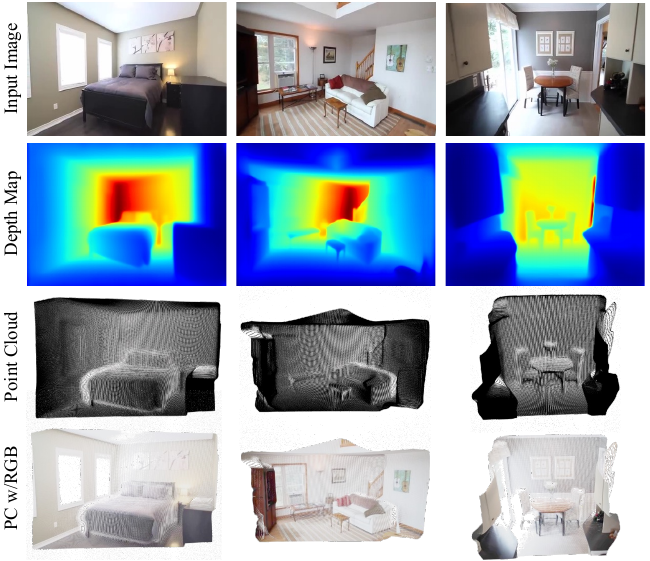}
    \vspace{-1.7em}
    \caption{
    Examples of input images with their corresponding estimated depth maps and back-projected 3D point clouds.
    For better visualization, we also present 3D point clouds with RGB colors.
    }
    \label{fig:depth_examples}
    \vspace{-1.8em}
\end{figure}

\subsection{Details on 3D Point Feature Extraction}
As described in Sec 3.3 in the main paper, we advocate incorporating 3D priors from 3D point features, which contain more comprehensive 3D domain knowledge than 2D depth maps, to address limited geometric information inherent in single-view settings.
In this section, we provide additional explanations on the procedure of producing 3D point features from a single source image.
As illustrated in Fig.~\ref{fig:3d_procedure}, our approach first extracts a pixel-wise depth map $D \in \mathbb{R}_{+}^{H \times W \times 1}$ from an input image $\mathcal{I} \in \mathbb{R}^{H \times W \times 3}$ using a pre-trained monocular depth estimation model~\cite{piccinelli2024unidepth}.
Next, we back-project $D$ into a 3D point cloud $P \in \mathbb{R}^{H \times W \times 3}$ with the corresponding camera parameters $K \in \mathbb{R}^{3 \times 3}$.
Then, a point cloud encoder takes $P$ to yield point features $F^S \in \mathbb{R}^{N_s \times D^S}$.
Here, we organize our point cloud encoder based on the prevalent PointNet~\cite{qi2017pointnet} architecture.
Given the points $P$, we sample $N_s$ points using the Farthest Point Sampling (FPS)~\cite{eldar1997farthest} algorithm; then, these sampled points $P^\prime \in \mathbb{R}^{N_s \times 3}$ are processed through a series of joint alignment networks and MLP layers.
The first alignment network maps the sampled points $P^\prime$ to a canonical space, and the second aligns intermediate features ${F^{\prime S}} \in \mathbb{R}^{N_s \times 64}$ to a joint feature space.
Both networks employ an affine transform matrix predicted by the T-Net.
Finally, we produce 3D point features $F^S \in \mathbb{R}^{N_s \times D^S}$, where $D^S$ denotes 1,024.
In Fig.~\ref{fig:depth_examples}, we present examples of input images $\mathcal{I}$, along with their corresponding depth maps $D$ and back-projected 3D point clouds $P$ (+ w/ RGB), to help understand our process.

\subsection{CATSplat Procedure}
In Algorithm.~\ref{alg:overview}, we present the overall workflow of our generalizable feed-forward network, incorporating two novel priors, for 3D scene reconstruction from a single image.

\SetKwComment{Comment}{\# }{}
\newcommand\mycommfont[1]{\small\textcolor{blue}{#1}}
\SetCommentSty{mycommfont}
\RestyleAlgo{ruled}
\setlength{\textfloatsep}{8pt}
\begin{algorithm}[h]
	\setlength{\belowcaptionskip}{-20pt}
	\caption{3D scene from a single-view image.}
	\label{alg:overview} 
        \KwIn{
        A monocular image $\mathcal{I} \in \mathbb{R}^{H \times W \times 3}$
        }
	\KwResult{
            Novel view images $\hat{\mathcal{I}}_t \in \mathbb{R}^{H \times W \times 3}$
            
            \textbf{Procedure:}
        }

        Estimate Depth Map $D$ from $\mathcal{I}$.

        Concatenate $\mathcal{I}$ and $D$ as $\mathcal{I}^\prime$.
        
        Extract multi-resolution image features $F_i^{\mathcal{I}}$ from $\mathcal{I}^\prime$.
        
        Produce text features $F_{i}^{C}$ based on the VLM.

        Back project $D$ into 3D points $P$.

        Produce 3D point features $F_{i}^{S}$ from $P$.
        
        \Comment{Multi-resolution Transformer with $N_l$ layers.}
        
        \For{$i = 1$ to $N_{l}$}{
            \Comment*[h]{Incorporation of Contextual Cues.}
            
            $\mathbf{Q}_i, \mathbf{K}_i, \mathbf{V}_i = W_q \cdot F_{i}^{\mathcal{I}},\;\: W_k \cdot F_{i}^{C},\;\:  W_v \cdot F_{i}^{C}$

            $F_i^{\mathcal{I}C} = Attn(\mathbf{Q}_i, \mathbf{K}_i, \mathbf{V}_i)$
            
            \Comment*[h]{Incorporation of Spatial Cues.}

            $\mathbf{Q}^\prime_i, \mathbf{K}^\prime_i, \mathbf{V}^\prime_i = W_q^\prime \cdot F_{i}^{\mathcal{I}C},\;\: W_k^\prime \cdot F_{i}^{S},\;\:  W_v^\prime \cdot F_{i}^{S}$

            $F_i^{\mathcal{I}CS} = Attn(\mathbf{Q}^\prime_i, \mathbf{K}^\prime_i, \mathbf{V}^\prime_i)$
            
            \Comment*[h]{Add and Normalization.}

            $\Tilde{F}_i^{\mathcal{I}CS} = \text{Norm}(F_{i}^{\mathcal{I}} + \bm{\gamma} \: \text{Dropout}(F_i^{\mathcal{I}CS}))$

            \Comment*[h]{Self Attention.}
            $\Tilde{\mathbf{Q}}_i, \Tilde{\mathbf{K}}_i, \Tilde{\mathbf{V}}_i = \Tilde{W_q} \cdot \Tilde{F}_i^{\mathcal{I}CS}, \Tilde{W_k} \cdot \Tilde{F}_i^{\mathcal{I}CS},  \Tilde{W_v} \cdot \Tilde{F}_i^{\mathcal{I}CS}$

            $\Tilde{F}_{i}^{\mathcal{I}} = Attn(\Tilde{\mathbf{Q}}_i, \Tilde{\mathbf{K}}_i, \Tilde{\mathbf{V}}_i)$

        }
        \Comment*[h]{3D Scene Reconstruction and Novel View Synthesis.}
        
        Predict $J$ Gaussians $\{(\bm{\mu}_j, \bm{\alpha}_j, \bm{\Sigma}_j, \bm{c}_j )\}^{J}_{j}$ from $\Tilde{F}_{i}^{\mathcal{I}}$.

        Render $\hat{\mathcal{I}}_t$ images with rasterization function.
\end{algorithm}
\section{Experimental Setup}
\label{supp:setup}

\subsection{Datasets}

\myparagraph{RealEstate10K.}
The RealEstate10K~\cite{zhou2018stereo} dataset consists of large-scale home walkthrough videos from YouTube, including approximately 10 million frames from around 80,000 videos.
It also provides camera parameters for each frame calibrated using the Structure-from-Motion (SfM) software. 
We follow the standard training and testing split, with 67,477 scenes for training and 7,289 for evaluation.

\myparagraph{NYUv2.}
The NYUv2~\cite{silberman2012indoor} dataset provides video sequences from diverse indoor environments captured using Kinect cameras.
In line with \cite{szymanowicz2024flash3d}, we employ 250 source images from 80 scenes for cross-dataset evaluation and randomly sample target frames within a $\pm$30 frame range from the source, following the random protocol of RE10K~\cite{zhou2018stereo}.
For camera trajectories, we use SfM software as RE10K.

\myparagraph{ACID.}
The ACID~\cite{liu2021infinite} dataset consists of large-scale natural landscape videos captured by aerial drones.
Like the RE10K~\cite{zhou2018stereo}, ACID provides camera parameters for frames, which are calculated via SfM software.
For cross-dataset evaluation, we utilize 450 source images from 150 scenes and randomly sample target frames within a $\pm$30 frame range from the source as the random protocol of RE10K.
Note that we evaluate and visualize Flash3D~\cite{szymanowicz2024flash3d} on ACID using publicly available code and provided checkpoints.

\myparagraph{KITTI.}
The KITTI~\cite{geiger2012we} is a landmark autonomous driving dataset containing 30 \textit{city} driving sequences.
Following the well-established evaluation protocol from \citet{tulsiani2018layer}, we utilize 1,079 source frames and provided corresponding camera parameters for cross-dataset evaluation.

\subsection{Scale Handling}
For training on the RealEstate10K~\cite{zhou2018stereo}, we employ the SfM~\cite{schonberger2016structure} strategy of COLMAP to estimate camera poses for images and align them with the estimated depth maps using the scale alignment process from Tucker~\etal~\cite{tucker2020single}.
While this approach is generally reliable, it does not assure perfect alignment between the reconstructed scene scale and novel view camera pose scale due to occasional outliers in the predictions.
Hence, when evaluating on RealEstate10K, ACID~\cite{liu2021infinite}, and NYUv2~\cite{silberman2012indoor}, we apply RANSAC to address such outliers, ensuring more robust scale handling.
For the KITTI~\cite{geiger2012we} dataset, we use the provided poses for evaluation.

\subsection{Implementation Details}
Our experimental setup is built on the prevalent deep learning framework, PyTorch.
For image processing, we utilize the ResNet-50~\cite{he2016deep} encoder and the UniDepth~\cite{piccinelli2024unidepth} pre-trained model for monocular depth estimation, with a single-view image size of $256 \times 384$.
We employ LLaVA~\cite{liu2024visual} 13B for text embeddings and extend the PointNet~\cite{qi2017pointnet} encoder for extracting point features. 
Note that we precompute text embeddings to optimize training efficiency by minimizing computational overhead.
Our multi-resolution transformer comprises three layers with 8-headed attention, leveraging three different resolution image features to effectively capture both global structures and fine details. 
We also set the ratio $\gamma$ as 0.5 to strike a balance, preventing excessive loss of core visual information from image features while integrating our two novel priors.
Then, our offset decoder predicts two sets of depth offsets and 3D offsets for vivid scene representation.
We train and evaluate on a single A100 GPU and select the best model upon convergence.
Specifically, we optimize a combination of $\mathcal{L}_{\ell1}$, $\mathcal{L}_{\textrm{ssim}}$, and $\mathcal{L}_{\textrm{lpips}}$ losses using the Adam optimizer with each coefficient as $\lambda_{\ell1}$=$1$, $\lambda_{\text{ssim}}$=$0.85$, and $\lambda_{\text{lpips}}$=$0.01$, respectively. 

\vspace{.1em}
\subsection{Computational Analysis}
In Tab.~\ref{tab:supp_costs}, we compare the training and inference times of ours with Flash3D~\cite{szymanowicz2024flash3d} on a single A100 GPU.
For precise examination, we synchronize CUDA events across all operations, as they are typically asynchronous per process.
Although CATSplat requires additional six hours (29.1 hrs) of training compared to Flash3D (23.2 hrs), the impact on inference time remains relatively minimal, increasing by only 0.066 secs.
In terms of model parameters, CATSplat (72M) is $30\%$ larger than Flash3D (54M) due to the transformer architecture, which introduces additional weight matrices and trainable layers.
Despite the slightly higher computational costs, CATSplat consistently outperforms Flash3D across multiple benchmarks by employing two valuable guidance.

\begin{table}[h]
\begin{center}
\vspace{-.5em}
\resizebox{\linewidth}{!}{%
\begin{tabular}{@{}lccccc@{}} \toprule
\multirow{2.3}{*}{Method} & \multirow{2.3}{*}{Training (hrs)} & \multirow{2.3}{*}{Inference (secs)} & \multicolumn{3}{c}{$n=\textit{Random}$ (frames)}\\\cmidrule{4-6}
     & & & PSNR $\uparrow$ & SSIM $\uparrow$ & LPIPS $\downarrow$\\\midrule
    Flash3D~\cite{szymanowicz2024flash3d} & 23.2 & 0.327 & 24.93 & 0.833 & 0.160\\
    CATSplat (Ours) & 29.1 & 0.393 & \textbf{25.45} & \textbf{0.841} & \textbf{0.151}\\
\bottomrule
\end{tabular}}
\vspace{-0.7em}
\caption{
Comparison of computational costs with the state-of-the-art single-view 3DGS method, Flash3D~\cite{szymanowicz2024flash3d}, on the RE10K~\cite{zhou2018stereo}.
}
\vspace{-2.8em}
\label{tab:supp_costs}
\end{center}
\end{table}
\section{Additional Experiments}
\label{supp:experiments}

\subsection{Ablation Studies in Cross-dataset Settings}
In this section, we validate the effectiveness of our two innovative priors through ablative experiments across cross-dataset settings.
In Tab.~\ref{tab:ablation_nyu} and Tab.~\ref{tab:ablation_acid}, we evaluate variants of our method, with/ and w/o Contextual and Spatial priors, on the NYUv2~\cite{silberman2012indoor} and ACID~\cite{liu2021infinite} datasets, respectively.
The Baseline denotes our basic transformer architecture, excluding cross-attention with any of our proposed priors.

\begin{table}[t]
\begin{center}
\vspace{-1.3em}
\resizebox{\linewidth}{!}{%
\begin{tabular}{@{}ccccccccc@{}} 
\toprule
\multicolumn{3}{c}{Method} & \multicolumn{3}{c}{$n=\textit{Random}$ (frames)}\\
\cmidrule{4-6}
Baseline & Contextual & Spatial \hspace{.3em} & PSNR $\uparrow$ & SSIM $\uparrow$ & LPIPS $\downarrow$\\ \midrule
$\checkmark$ & - & - & 25.11 & 0.775 & 0.178\\
$\checkmark$ & $\checkmark$ & - & 25.51 & 0.779 & 0.163\\
$\checkmark$ & - & $\checkmark$ & 25.48 & 0.778 & 0.165\\
$\checkmark$ & $\checkmark$ & $\checkmark$ & \textbf{25.57} & \textbf{0.781} & \textbf{0.157}\\
\bottomrule
\end{tabular}}
\vspace{-0.5em}
\caption{
Ablation study to investigate the effect of our two intelligent priors on the NYUv2~\cite{silberman2012indoor} dataset in cross-dataset settings.
}
\vspace{-1.5em}
\label{tab:ablation_nyu}
\end{center}
\end{table}

\begin{table}[t]
\begin{center}
\resizebox{\linewidth}{!}{%
\begin{tabular}{@{}ccccccccc@{}} 
\toprule
\multicolumn{3}{c}{Method} & \multicolumn{3}{c}{$n=\textit{Random}$ (frames)}\\
\cmidrule{4-6}
Baseline & Contextual & Spatial \hspace{.3em} & PSNR $\uparrow$ & SSIM $\uparrow$ & LPIPS $\downarrow$\\ \midrule
$\checkmark$ & - & - & 24.26 & 0.732 & 0.261\\
$\checkmark$ & $\checkmark$ & - & 24.57 & 0.735 & 0.253\\
$\checkmark$ & - & $\checkmark$ & 24.62 & 0.737 & 0.254\\
$\checkmark$ & $\checkmark$ & $\checkmark$ & \textbf{24.73} & \textbf{0.739} & \textbf{0.250}\\
\bottomrule
\end{tabular}}
\vspace{-0.5em}
\caption{
Ablation study to investigate the effect of our two intelligent priors on the ACID~\cite{zhou2018stereo} dataset in cross-dataset settings.
}
\vspace{-1.em}
\label{tab:ablation_acid}
\end{center}
\end{table}

First, incorporating contextual cues leads to significant improvements, both for indoor scenes (NYUv2) and outdoor nature scenes (ACID).
With text embeddings from a well-trained visual-langualge model (VLM)~\cite{liu2024visual}, our network learns not just basic object types or scene semantics but also deeper context, such as how objects relate to each other or the overall scene structure.
In other words, we take advantage of text embeddings to provide comprehensive general knowledge of dynamic real-world environments as well as scene-specific details.
As a result, even in unfamiliar scenarios without familiar scene types or objects, text-embedded general cues serve as guiding anchors, enabling our network to better apply trained knowledge.

Additionally, by incorporating spatial guidance, our approach boosts generalization performance on both datasets.
Beyond the geometric cues from 2D depth maps, we guide our network to be aware of three-dimensional domains, more associated with 3D Gaussians, through 3D point features.
Based on deep spatial understandings, our network effectively reconstructs 3D scenes with accurate Gaussians, even in complex, unfamiliar environments.
Finally, combining all priors together achieves further advances, seamlessly complementing limited knowledge from single-view image features.
In addition to Tab.4 in our main paper, these results demonstrate the significance of our two novel priors.

\begin{figure}[t]
    \vspace{-1.7em}
    \centering
    \includegraphics[width=\linewidth]{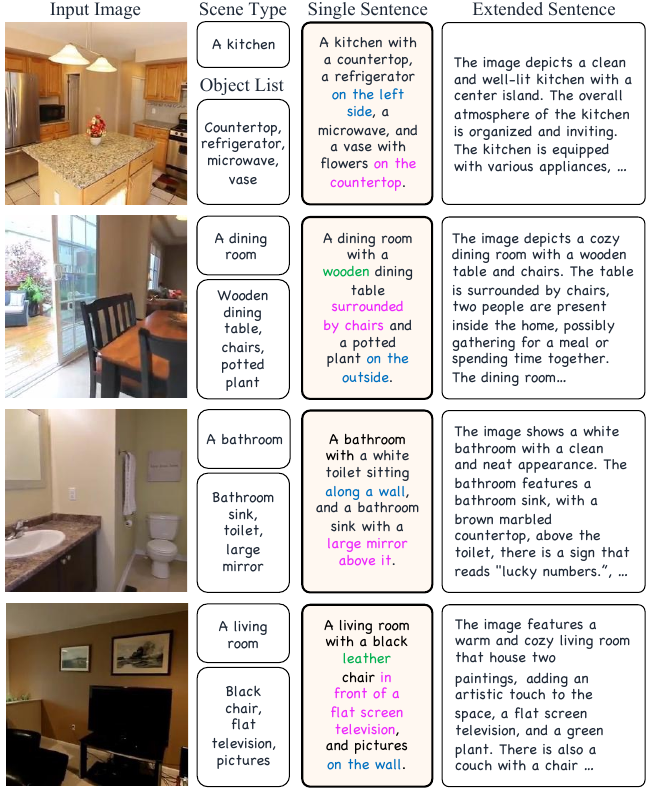}
    \vspace{-1.7em}
    \caption{
    Examples of four different formats of text descriptions from the VLM~\cite{liu2024visual}, as described in Tab.5 in the main paper.
    }
    \label{fig:textdescriptions}
\end{figure}

\begin{figure*}[t]
    \vspace{-1.5em}
    \centering
    \includegraphics[width=\linewidth]{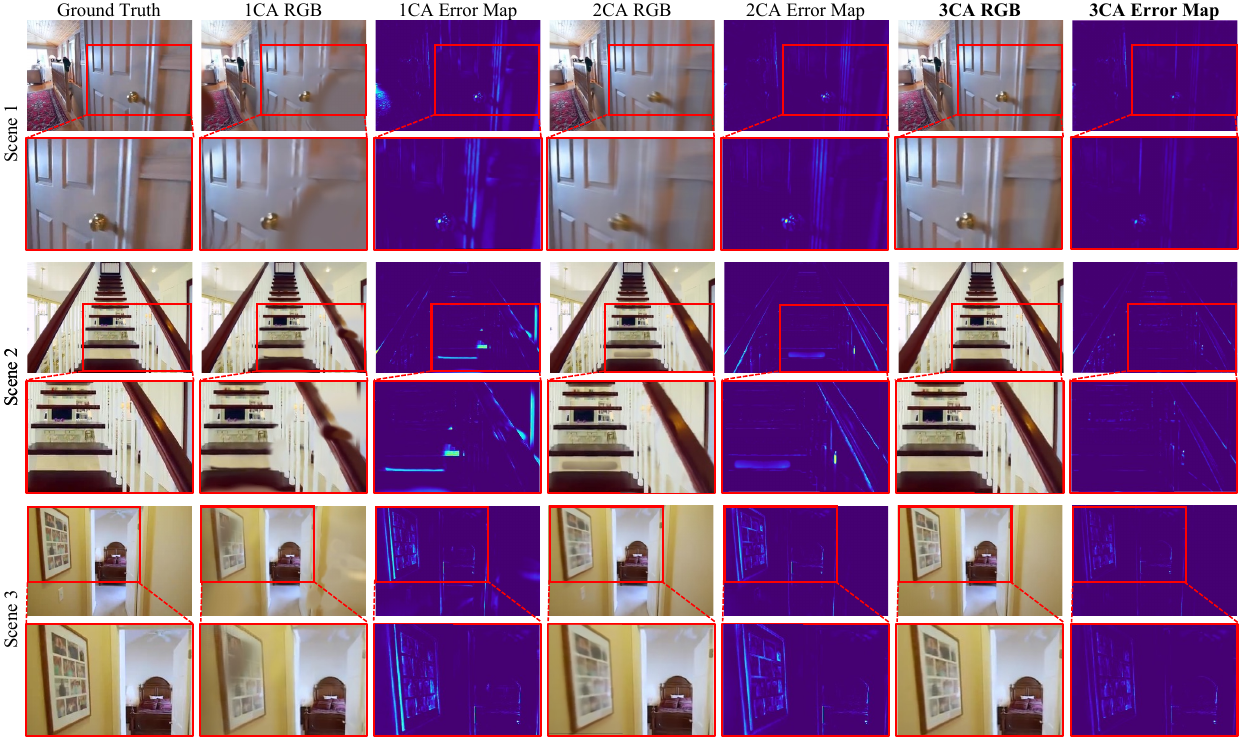}
    \vspace{-1.5em}
    \caption{
    Ablation study to see the effect of iteratively incorporating our novel priors on the RE10K~\cite{zhou2018stereo} ($n$=\textit{Random}).
    For clear ablations, we keep the number of entire transformer layers consistent across the experiments and adjust only the number of cross-attentions (CA).
    }
    \label{fig:supp_ca}
    \vspace{-1.3em}
\end{figure*}

\subsection{Discussion on Text Descriptions}
For rich contextual cues, we leverage text embeddings from a well-trained VLM~\cite{liu2024visual}.
Specifically, we prompt the VLM to generate text descriptions for the input image; then, we utilize intermediate text embeddings before they are processed into linguistic descriptions.
To discover the optimal text embeddings, we investigate the impact of contextual information within different types of text embeddings on generalizability in Tab.5 (main).
For comparison, we conduct experiments with four different styles of prompts: identifying the scene type, listing objects, describing the scene with a detailed single sentence, and two or more extended sentences.
We provide examples of text description outputs using these prompts in Fig.~\ref{fig:textdescriptions}.
Usually, a single sentence captures comprehensive details for the scene, including textures (\eg, \textit{``wooden''}, \textit{``leather''}), object relationships (\eg, \textit{``on the countertop''}, \textit{``surrounded by chairs''}, \textit{``large mirror above it''}), and overall composition (\eg, \textit{``on the left side''}, \textit{``on the outside''}), surpassing simple cues like scene type or object list.
However, extended sentences often introduce exaggerated or fabricated elements, such as overly interpretive moods, atmospheric descriptions with excessive adjectives (\eg, \textit{``organized and inviting''}, \textit{``adding an artistic touch''}), or entirely false specifics (\eg, \textit{``two people are present inside the home..''}, \textit{``lucky numbers..''}).
These noisy overstatements hinder the network from learning meaningful context information of the text embeddings, resulting in relatively lower performance than using a single sentence.
Ultimately, in this work, we employ a single sentence to enhance image features with practical contextual information.

\subsection{Text Embeddings from Various VLMs}
Contextual cues from text embeddings are one of our core methods to break through the inherent constraints in monocular settings.
Thus, identifying the most effective text embeddings is crucial for achieving high-quality single-view 3D scene reconstruction.
In Tab.~\ref{tab:supp_vlm}, we explore how text embeddings from various pre-trained VLMs, including OpenFlamingo~\cite{awadalla2023openflamingo}, BLIP2 T5~\cite{li2023blip}, LLaVA 7B~\cite{liu2024visual}, and LLaVA 13B, influence performance on the RE10K~\cite{zhou2018stereo}.
For a fair comparison, we prompt all VLMs to produce a single sentence scene description.
Then, we use intermediate text embeddings from each VLM.
Even with similar prompts, each model generates distinct structures of text descriptions.
For example, OpenFlamingo tends to produce relatively unstable text descriptions with redundant or exaggerated information, providing minimal benefit for scene reconstruction.
Meanwhile, BLIP2 and LLaVA 7B generate monotonous text descriptions that primarily focus on object and scene types.
On the other hand, LLaVA 13B yields more informative text descriptions with useful details, such as textures (\eg, \textit{``wooden''}, \textit{``leather''}), object relations (\eg, \textit{``on the countertop''}, \textit{``surrounded by chairs''}, \textit{``large mirror above it''}), and scene composition (\eg, \textit{``on the left side''}, \textit{``on the outside''}), as illustrated in Fig.~\ref{fig:textdescriptions}.
Finally, we leverage text embeddings from the well-aligned multimodal space of LLaVA 13B for context-aware 3D scene reconstruction.

\begin{table}[t]
\begin{center}
\vspace{.3em}
\resizebox{\linewidth}{!}{%
\begin{tabular}{@{}lcccccc@{}} \toprule
& \multicolumn{3}{c}{$n=\text{10}$ (frames)} & \multicolumn{3}{c}{$n=\textit{Random}$ (frames)} \\\cmidrule{2-4} \cmidrule{5-7}
    Method & PSNR $\uparrow$ & SSIM $\uparrow$ & LPIPS $\downarrow$ & PSNR $\uparrow$ & SSIM $\uparrow$ & LPIPS $\downarrow$\\\midrule
    OpenFlamingo & 26.08 & 0.858 & 0.131 & 25.06 & 0.832 & 0.158\\
    BLIP2 T5 & 26.29 & 0.860 & 0.129 & 25.27 & 0.833 & 0.156\\
    LLaVA 7B & 26.19 & 0.861 & 0.129 & 25.23 & 0.834 & 0.156\\
    LLaVA 13B & \textbf{26.40} & \textbf{0.864} & \textbf{0.127} & \textbf{25.40} & \textbf{0.838} & \textbf{0.153}\\
\bottomrule
\end{tabular}}
\vspace{-0.7em}
\caption{
Ablation study to see the impact of text embeddings from different VLMs: OpenFlamingo~\cite{awadalla2023openflamingo}, BLIP2~\cite{li2023blip}, and LLaVA~\cite{liu2024visual}.
}
\vspace{-1.2em}
\label{tab:supp_vlm}
\end{center}
\end{table}

\begin{figure*}[t]
    \vspace{-1.4em}
    \centering
    \includegraphics[width=\linewidth]{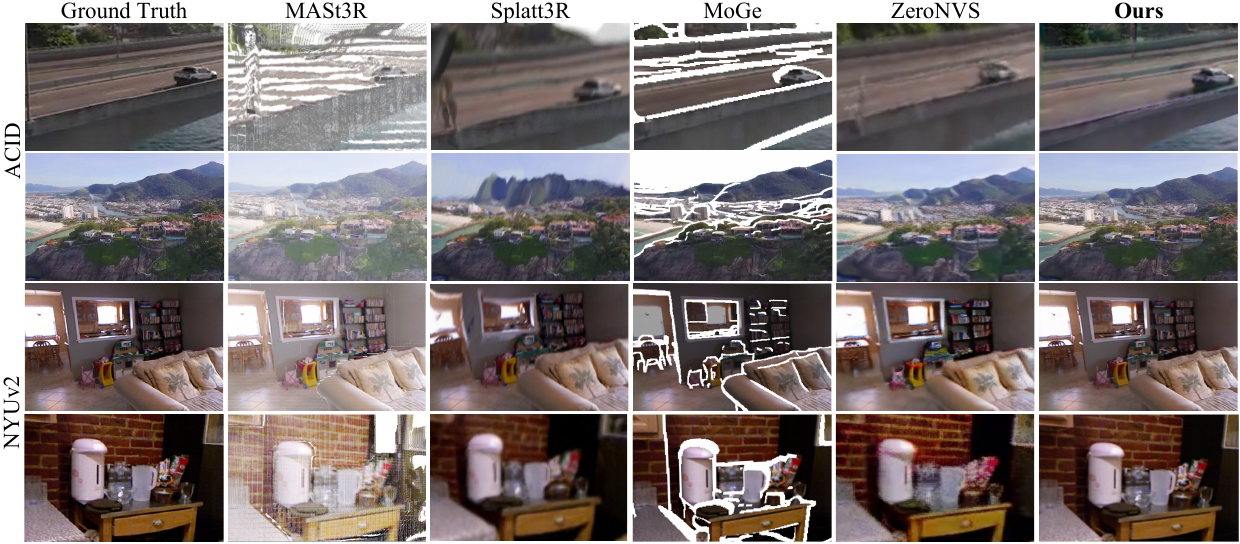}
    \vspace{-1.7em}
    \caption{
    Qualitative comparisons with various novel view synthesis approaches, including MASt3R~\cite{leroy2024grounding}, Splatt3R~\cite{smart2024splatt3r}, MoGe~\cite{wang2024moge}, and ZeroNVS~\cite{sargent2024zeronvs}.
    All methods are fairly evaluated on unseen scenes in cross-dataset settings across ACID~\cite{liu2021infinite} and NYUv2~\cite{silberman2012indoor} datasets.
    }
    \label{fig:supp_additional}
    \vspace{-1.em}
\end{figure*}

\vspace{-.1em}
\subsection{Iteratively Incorporating Priors}
We present additional ablative experimental results to highlight the benefits of iteratively incorporating our priors in Fig.~\ref{fig:supp_ca}.
Consistent with the settings in Fig.5 (main), we randomly sample the target frame within a $\pm$30 range; also, fix the total number of transformer layers at three and apply cross-attention either in the first layer only, across two layers, or throughout all three layers.
Through iterative cross-attention between image features and our priors, blurry artifacts gradually fade, sharpening the object contours and enhancing clarity in images.
Simultaneously, errors between rendered images and target images also steadily decrease.
In essence, iterative incorporations of our novel priors lead to noticeable improvements in overall visual quality.
These results emphasize both the importance of our priors and the structural robustness of our transformer-based architecture.

\begin{figure}[t]
    \centering
    \includegraphics[width=\linewidth]{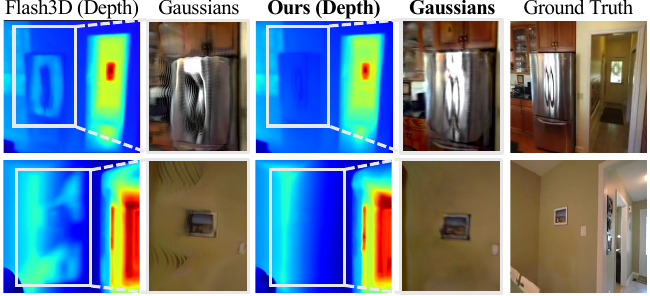}
    \vspace{-1.8em}
    \caption{
    Qualitative comparisons of 3D reconstruction between Flash3D~\cite{szymanowicz2024flash3d} and ours with Ground Truth.
    We visualize zoom-in views of 3D Gaussian distributions and depth maps from them.
    }
    \label{fig:depth}
\end{figure}

\subsection{Comparisons with Various Approaches}
In addition to comparing our method with monocular novel view synthesis (NVS) approaches, we evaluate ours against various types of NVS methods to further highlight its effectiveness.
For a fair comparison, we assess all methods on unseen scenes in cross-dataset settings across ACID~\cite{liu2021infinite} and NYUv2~\cite{silberman2012indoor} datasets.
First, we compare ours with MASt3R~\cite{leroy2024grounding} and Splatt3R~\cite{smart2024splatt3r}, both optimized for processing stereo image pairs in 3D scene reconstruction.
MASt3R relies on feature-based matching between image pairs, and Splatt3R extends MASt3R by incorporating additional prediction heads to directly estimate Gaussian primitives.
While both methods excel in 3D reconstruction with two images, they tend to produce relatively blurry results in monocular settings due to their inherent dependence on dual inputs.
MoGe~\cite{wang2024moge} extracts high-level features from a single image and regresses these features into 3D point maps for 3D scene reconstruction.
However, as shown in the fourth column, unlike ours, which represents the 3D scene using continuous 3D Gaussians, the use of 3D points often results in large gaps in the outlines of objects from novel viewpoints.
Moreover, ZeroNVS incorporates~\cite{sargent2024zeronvs} diffusion processes based on NeRF architecture to construct detailed 3D scenes from a single image.
Although it achieves high-quality novel view images, combining diffusion with NeRF-based volume rendering demands significantly higher computational costs ($\approx$ 3 hrs).
On the other hand, CATSplat efficiently reconstructs solid scenes for novel viewpoints using explicit 3D Gaussian distributions in a single forward pass.


\subsection{Additional Comparisons with Flash3D}
In addition to comparing rendered RGBs, we qualitatively assess the quality of 3D Gaussians for scene representation.
In Fig.~\ref{fig:depth}, ours predicts clearer Gaussians than Flash3D~\cite{szymanowicz2024flash3d}, which exhibits messy artifacts.
Our excellence is also evident in the depth maps produced by these 3D Gaussians.
We present more qualitative comparisons with Flash3D on the RE10K~\cite{zhou2018stereo} (Fig.~\ref{fig:supp_re10k_1} and Fig.~\ref{fig:supp_re10k_2}) as well as ACID~\cite{liu2021infinite} (Fig.~\ref{fig:supp_acid}) and KITTI~\cite{geiger2012we} (Fig.~\ref{fig:supp_kitti}) in cross-dataset settings.



\section{Limitations and Future Work}
\label{supp:limitations}
\begin{figure}[t]
    \centering
    \includegraphics[width=\linewidth]{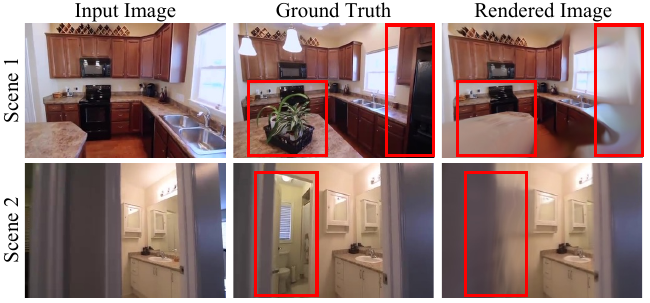}
    \vspace{-1.9em}
    \caption{
    Failure cases of CATSplat.
    When invisible areas in the input become visible in the target, ours might be less productive.
    }
    \label{fig:failure}
\end{figure}

Although CATSplat shines in monocular 3D scene reconstruction with two additional priors, it does not ensure perfect novel view synthesis across all real-world scenarios. 
Depending on dynamic camera movements, when regions that are occluded, truncated, or even entirely missing in the input image appear in the target view, ours might be less effective.
For example, in Fig.~\ref{fig:failure}, when previously unseen elements, like green plants absent in the input, emerge in the target view (Scene1) or when areas of the bathroom, once hidden behind a door, become visible (Scene2), our model struggles to reconstruct these newly revealed parts.
In the future, we plan to explore involving generative knowledge to better handle these unseen regions in monocular 3D scene reconstruction.
Moreover, we believe that training the model on a broader range of datasets will strengthen its general understanding of challenging natural environments.

\begin{figure*}[t]
    \vspace{-1em}
    \centering
    \includegraphics[width=0.97\linewidth]{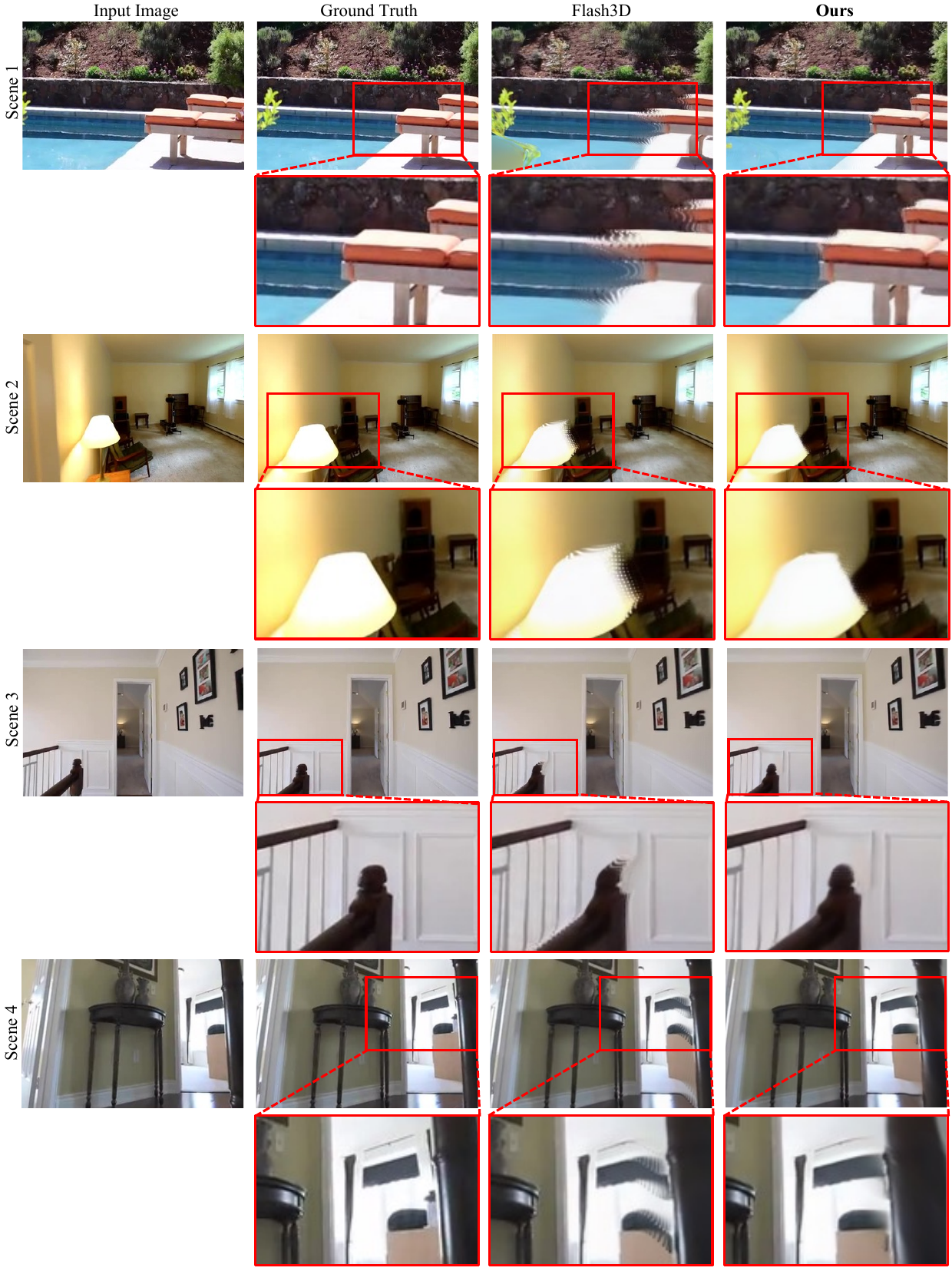}
    \caption{
    Qualitative comparisons between Flash3D~\cite{szymanowicz2024flash3d} and Ours with Input Image and Ground Truth on the RealEstate10K~\cite{zhou2018stereo} dataset.
    }
    \label{fig:supp_re10k_1}
\end{figure*}

\begin{figure*}[t]
    \vspace{-1em}
    \centering
    \includegraphics[width=.97\linewidth]{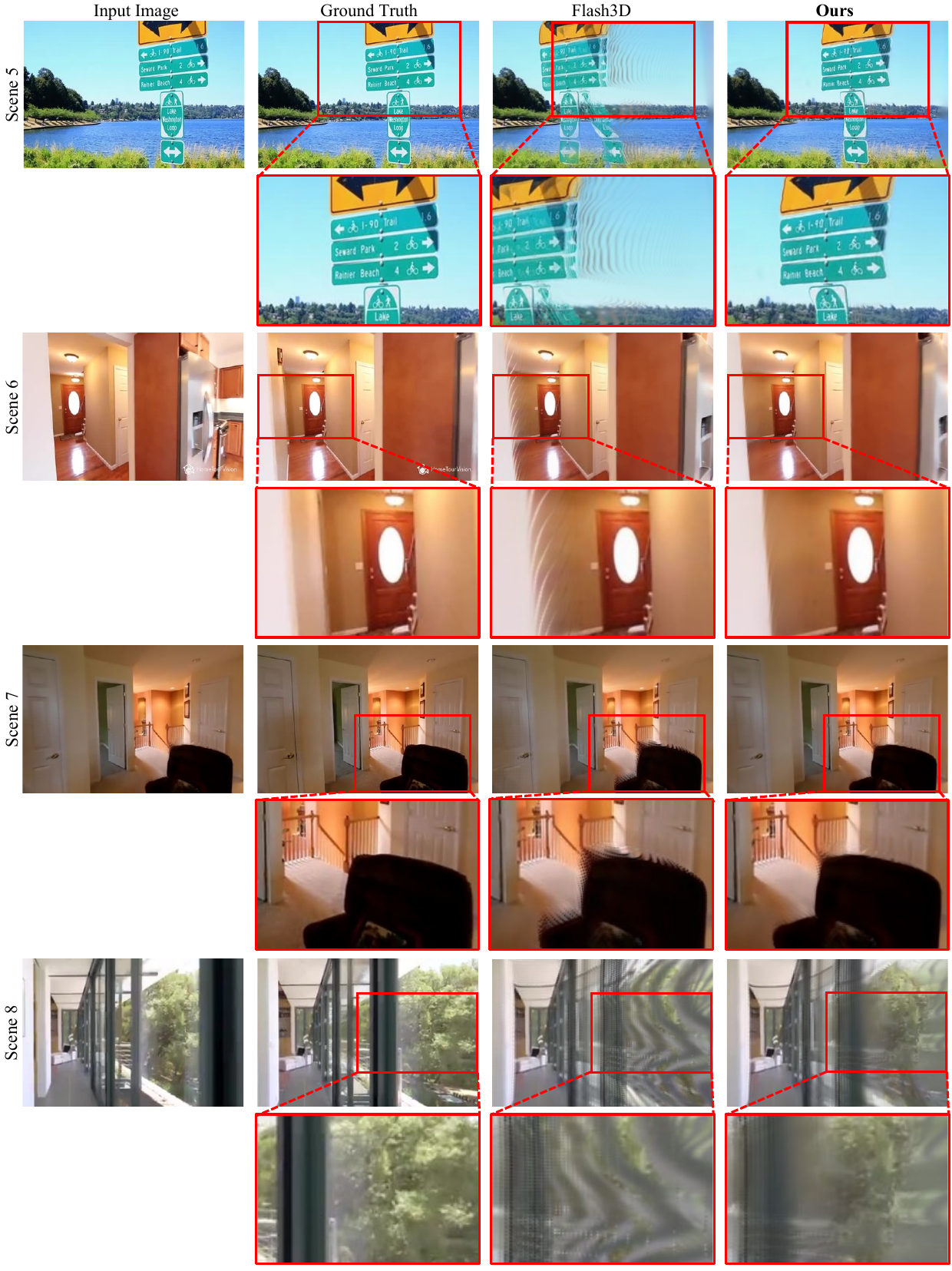}
    \caption{
    Qualitative comparisons between Flash3D~\cite{szymanowicz2024flash3d} and Ours with Input Image and Ground Truth on the RealEstate10K~\cite{zhou2018stereo} dataset.
    }
    \label{fig:supp_re10k_2}
\end{figure*}

\begin{figure*}[t]
    \vspace{-1em}
    \centering
    \includegraphics[width=.97\linewidth]{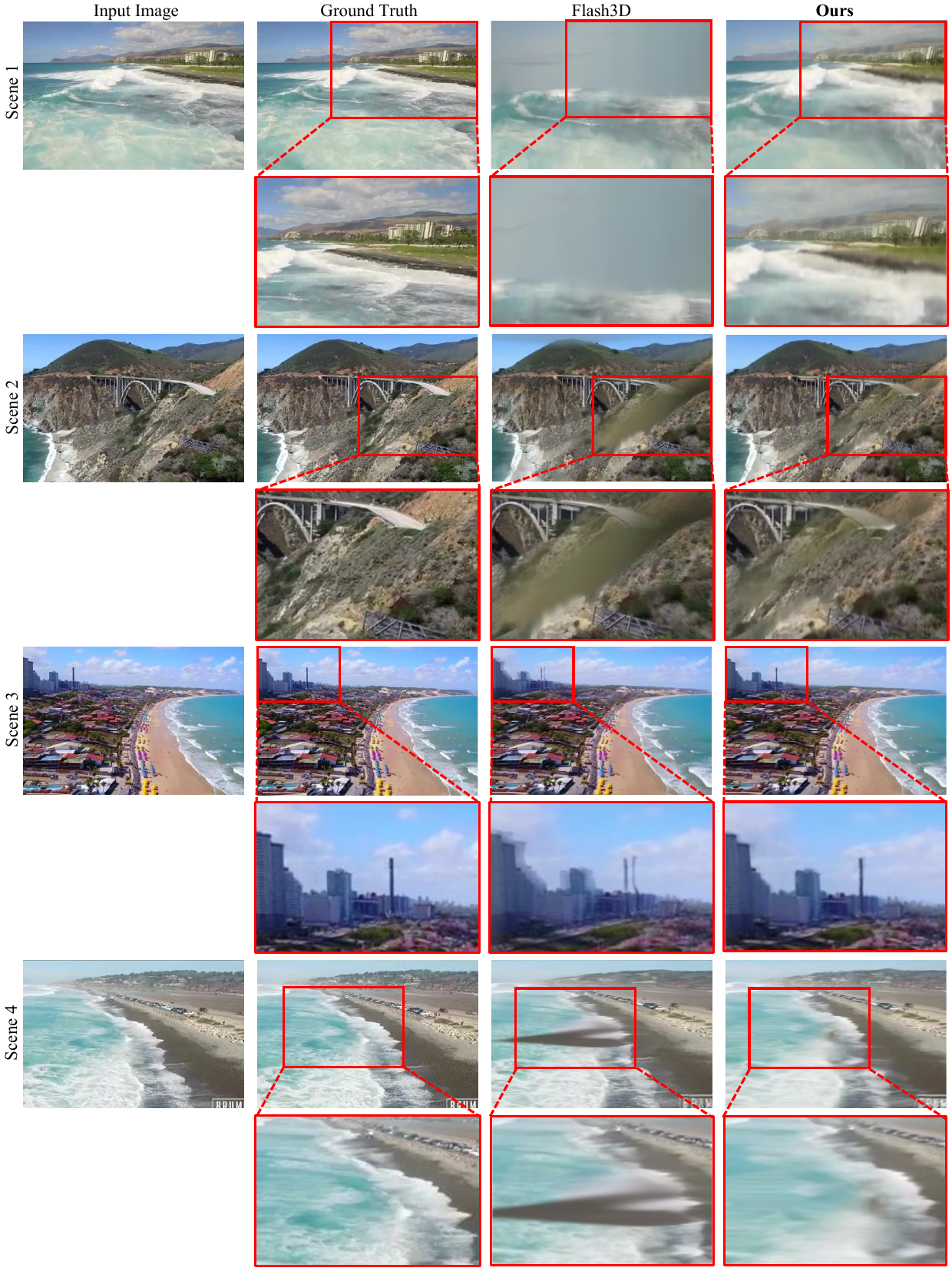}
    \caption{
    Qualitative comparisons between Flash3D~\cite{szymanowicz2024flash3d} and Ours with Input Image and Ground Truth on the ACID~\cite{liu2021infinite} dataset.
    }
    \label{fig:supp_acid}
\end{figure*}

\begin{figure*}[t]
    \vspace{-1em}
    \centering
    \includegraphics[width=.97\linewidth]{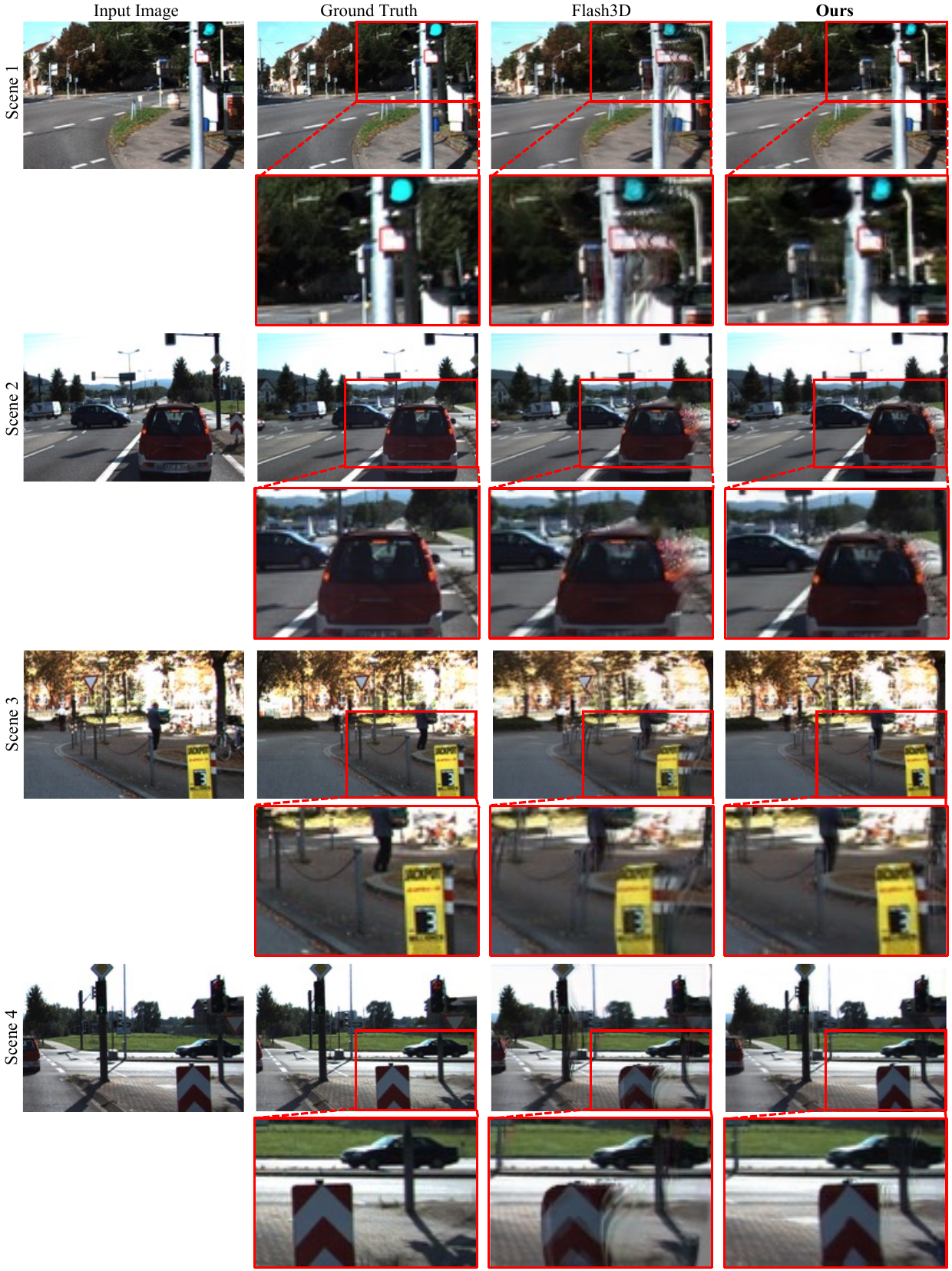}
    \caption{
    Qualitative comparisons between Flash3D~\cite{szymanowicz2024flash3d} and Ours with Input Image and Ground Truth on the KITTI~\cite{geiger2012we} dataset.
    }
    \label{fig:supp_kitti}
\end{figure*}

\clearpage
{
    \small
    \bibliographystyle{ieeenat_fullname}
    \bibliography{main}
}